\documentclass[sigconf]{acmart}
\usepackage{amsmath,amsfonts}
\usepackage{algorithmic}
\usepackage{graphicx}
\usepackage{textcomp}
\usepackage{xcolor}
\usepackage{todonotes}
\usepackage{url}
\usepackage{soul}
\usepackage{xspace}
\usepackage{booktabs}
\usepackage{multirow}
\usepackage{framed}
\usepackage{subcaption}
\usepackage{enumitem}
\usepackage{comment}
\usepackage{fancybox}
\usepackage{soul}

\newcommand{\jack}[1]{\textcolor{purple}{{\it [Jack says: #1]}}}

\newcommand{\huawei}{\textsf{Huawei}\xspace}

\newcommand{\hypobox}[1]{\begin{center}%	
		\noindent\thicklines\setlength{\fboxsep}{7pt}%	
		\cornersize{0}\Ovalbox{\begin{minipage}{0.9\columnwidth}%	
				\vspace{-0.1cm}
				\textit{#1}
				\vspace{-0.1cm}
\end{minipage}} \end{center}}

\AtBeginDocument{%
  \providecommand\BibTeX{{%
    \normalfont B\kern-0.5em{\scshape i\kern-0.25em b}\kern-0.8em\TeX}}}

\copyrightyear{2022} 
\acmYear{2022} 
\setcopyright{acmcopyright}\acmConference[ICSE '22]{44th International Conference on Software Engineering}{May 21--29, 2022}{Pittsburgh, PA, USA}
\acmBooktitle{44th International Conference on Software Engineering (ICSE '22), May 21--29, 2022, Pittsburgh, PA, USA}
\acmPrice{15.00}
\acmDOI{10.1145/3510003.3510163}
\acmISBN{978-1-4503-9221-1/22/05}

\begin{document}

%\title{A Systematic Approach towards Reproducible Deep Learning Experiments}
\title{Towards Training Reproducible Deep Learning Models}

\author{Boyuan Chen}
\affiliation{%
	\institution{Centre for Software Excellence, Huawei Canada}
	\city{Kingston}
	\country{Canada}}
\email{boyuan.chen1@huawei.com}

\author{Mingzhi Wen}
\affiliation{%
	\institution{Huawei Technologies}
	\city{Shenzhen}
	\country{China}}
\email{wenmingzhi@huawei.com}

\author{Yong Shi}
\affiliation{%
	\institution{Huawei Technologies}
	\city{Shenzhen}
	\country{China}}
\email{young.shi@huawei.com}

\author{Dayi Lin}
\affiliation{%
	\institution{Centre for Software Excellence, Huawei Canada}
	\city{Kingston}
	\country{Canada}}
\email{dayi.lin@huawei.com}

\author{Gopi Krishnan Rajbahadur}
\affiliation{%
	\institution{Centre for Software Excellence, Huawei Canada}
	\city{Kingston}
	\country{Canada}}
\email{gopi.krishnan.rajbahadur1@huawei.com}

\author{Zhen Ming (Jack) Jiang}
\affiliation{%
	\institution{York University}
	\city{Toronto}
	\country{Canada}
}
\email{zmjiang@eecs.yorku.ca}

\renewcommand{\shortauthors}{Chen and Wen, et al.}

\begin{abstract}

Reproducibility is an increasing concern in Artificial Intelligence (AI), particularly in the area of Deep Learning (DL). Being able to reproduce DL models is crucial for AI-based systems, as it is closely tied to various tasks like training, testing, debugging, and auditing. However, DL models are challenging to be reproduced due to issues like randomness in the software (e.g., DL algorithms) and non-determinism in the hardware (e.g., GPU). There are various practices to mitigate some of the aforementioned issues. However, many of them are either too intrusive or can only work for a specific usage context.
In this paper, we propose a systematic approach to training reproducible DL models. Our approach includes three main parts: (1) a set of general criteria to thoroughly evaluate the reproducibility of DL models for two different domains, (2) a unified framework which leverages a record-and-replay technique to mitigate software-related randomness and a profile-and-patch technique to control hardware-related non-determinism, and (3) a reproducibility guideline which explains the rationales and the mitigation strategies on conducting a reproducible training process for DL models. Case study results show our approach can successfully reproduce six open source and one commercial DL models.

\end{abstract}


%\ccsdesc[500]{Software and its engineering~Software development methods}
%\begin{CCSXML}
%<ccs2012>
%  <concept_id>10010520.10010553.10010562</concept_id>
%  <concept_desc>Computer systems organization~Embedded systems</concept_desc>
%  <concept_significance>500</concept_significance>
%  <concept_id>10010520.10010575.10010755</concept_id>
%  <concept_desc>Computer systems organization~Redundancy</concept_desc>
%  <concept_significance>300</concept_significance>
%  <concept_id>10010520.10010553.10010554</concept_id>
%  <concept_desc>Computer systems organization~Robotics</concept_desc>
%  <concept_significance>100</concept_significance>
%  <concept_id>10003033.10003083.10003095</concept_id>
%  <concept_desc>Networks~Network reliability</concept_desc>
%  <concept_significance>100</concept_significance>
%</ccs2012>
%\end{CCSXML}
%
%\ccsdesc[500]{Computer systems organization~Embedded systems}



\maketitle

\section{Introduction}
\label{sec:intro}

In recent years, Artificial Intelligence (AI) has been advancing rapidly both in research and practice.
A recent report by McKinsey estimates that AI-based applications have the potential market values ranging from \$3.5 and \$5.8 trillion annually~\cite{mckinseyreport}. Many of these applications, which can perform complex tasks such as autonomous driving~\cite{GrigorescuJFR20}, speech recognition~\cite{AmodeiICML16}, and healthcare~\cite{esteva2019guide}, are enabled by various Deep Learning (DL) models~\cite{LeCunNature15}. 
Unlike traditional software systems, which are programmed based on deterministic rules (e.g., if/else), the DL models within AI-based systems are constructed in a stochastic way due to the underlying DL algorithms, whose behavior may not be reproducible and trustworthy~\cite{BrundageArxiv20,ParnasACM17}.
Ensuring the reproducibility of DL models is vital for not only many product development related tasks such as training~\cite{LiuTOSEM20}, testing~\cite{testMLGoogle}, debugging~\cite{PhamASE20} and legal compliance~\cite{assessmentEurope}, but also facilitating scientific movements like open science~\cite{woelfle2011open,vicente2018open}.
%Hence, ensuring DL based systems to be deterministic is important for various tasks (e.g., , diagnosis, debugging, and testing).
%Ensuring reproducibility of DL experiments improves the transparency of research and helps researchers to build on existing works and make further scientific contributions~\cite{woelfle2011open}. It also plays a key role in the open science movement~\cite{woelfle2011open,vicente2018open}.
%In practice, a reproducible DL-based software system will contribute to auditing and legal compliance, increase the trustworthiness of the product, and gain trust among customers. 

One of the important steps towards reproducible AI-based systems is to ensure the reproducibility of the DL models during the training process. A DL model is \emph{reproducible}, if under the same training setup (e.g., the same training code, the same environment, and the same training dataset), the resulting trained DL model yields the same results under the same evaluation criteria (e.g., the same evaluation metrics on the same testing dataset)~\cite{pineau2020improving,PhamASE20}. 
%DL models are reproducible, if identical training process yield identical results~\cite{pineau2020improving,PhamASE20}.
%The identical training process includes the same training code, configurations, and datasets. The examples of evaluation metrics for results are accuracy (for classification tasks) and loss values (for regression tasks).
%\jack{P2: start saying ML model reproduciabiity is hard, cite reproduciablity crisis paper and cite Xin'a paper on SE4AI reproduciability. Then briefly explain the three main sources of non-determinisms causing non-reproduciability and their current state and challenges: }
Unfortunately, recent studies show that AI faces reproducibility crisis~\cite{HutsonScience18,GundersenAAAI18}, especially for DL models~\cite{LiuTOSEM20,PhamASE20,LeeRML19,tatman2018practical,RaffNIPS19,sugimura2018building,isdahl2019out,ghanta2018systems}. In general, there are three main challenges associated with this: 

\begin{itemize}[leftmargin=*]

	%DL is a rapidly growing field and the relevant assets in the experiment are quickly evolving. Such changed need to be version managed to avoid non-deterministic behavior of DL models. For example, using different software dependencies (e.g., different DL frameworks) will cause the trained DL models to behave differently~\cite{WuArxiv18, PhamASE20}. Even with the same dependent software package, version difference might also have an impact on the experiment results. Hence, it is important to ensure that the original experiment and the reproducing experiment have identical assets prior to the training process. 
	
	%\noindent 
	\item \textbf{Randomness in the software~\cite{ScardapaneWI17}}: Randomness is essential in DL model training like batch ordering, data shuffling, and weight initialization for constructing robust and high-performing DL models ~\cite{PhamASE20,pytorchrandom}. However, randomness prevents the DL models from being reproduced. To achieve reproducibility in the training process, the current approach is to set predefined random seeds before the training process. Although this approach is effective in controlling the randomness, it has three drawbacks: (1) it might cause the training process to converge to local optimums and not able to explore other optimization opportunities; (2) it is non-trivial to select the appropriate seeds as there are no existing techniques for tuning random seeds during the hyperparameter tuning process; (3) non-trivial manual efforts are needed to locate randomness introducing functions and instrument them with seeds for the imported libraries and their dependencies.%It also helps to build a more robust and generic models. The randomness in software will cause the DL models to behave non-deterministically and needs to be mitigated. 
	
	%\noindent 
	\item \textbf{Non-determinism in the hardware~\cite{duncanGTC19}}: Training DL models requires intensive computing resources. For example, many matrix operations occur in the backward propagation, which consists of a huge amount of floating point operations.
	%Using CPUs for training is mostly used for models with simpler structures or quickly verifying ideas. 
	As GPUs have way more numbers of cores than CPUs, GPUs are often used for running DL training processes due to their ability to process multiple operations in parallel.
	However, executing floating point calculation in parallel becomes a source of non-determinism since the results of floating-point operations are sensitive to computation orders due to rounding errors~\cite{PhamASE20,GoldbergCS91}. 
	In addition, GPU specific libraries (e.g., CUDA~\cite{cuda} and cuDNN~\cite{ChetlurArxiv14}) by default auto-select the optimal primitive operations based on comparing different algorithms of operations during runtime (i.e., the auto-tuning feature). However, the comparison results might be non-deterministic due to issues like floating point computation mentioned above~\cite{pytorchrandom,PhamASE20}.
	These sources of non-determinism from hardware need to be controlled in order to construct reproducible DL models. Case-by-case solutions have been proposed to tackle specific issues. For example, both Pytorch~\cite{PaszkeNIPS19} and TensorFlow~\cite{AbadiOSDI16} provide configurations on disabling the auto-tuning feature. Unfortunately, none of these techniques have been empirically validated in literature. Furthermore, there is still a lack of a general technique which can work across different software frameworks.
	%Instead, practitioners leverage GPUs or specially designed TPUs for this task.
	% One of the most common approach is to leverage the CUDA and CuDNN library to interact with these types of hardware. However, some operations are non-deterministic due to the floating point precisions.
	
	%\noindent 
	\item \textbf{Lack of systematic guidelines~\cite{PhamASE20}}:
	Various checklists and documentation frameworks~\cite{mlreprochecklist,GebruDatasheet18,MitchellFAT19} have been proposed on asset management to support DL reproducibility. There are generally four types of assets to manage in machine learning (ML) in order to achieve model reproducibility: resources (e.g., dataset and environment), software (e.g., source code), metadata (e.g., dependencies), and execution data (e.g., execution results)~\cite{IdowuSEIP21}. %These assets need to be carefully managed for reproducibility. 
	However, prior work~\cite{IdowuSEIP21,AmershiSEIP19,BarrakSANER21} shows these assets should not be managed with the same toolsets (e.g., Git) used for source code~\cite{AmershiSEIP19}. Hence, new version management tools (e.g., DVC~\cite{dvc} and MLflow~\cite{mlflow}) are specifically designed for managing ML assets.
	However, even by adopting the techniques and suggestions mentioned above, DL models cannot be fully reproduced~\cite{PhamASE20} due to problems mentioned in the above two challenges. A systematic guideline is needed for both researchers and practitioners in order to construct reproducible DL models.
\end{itemize}

% [P4] We have found there are mainly three sources of non-determinism that could impact the reproducibility of DL experiments: (a) the software level (b) the hardware level (c) the environment level.

%\jack{P3: explain current state and their problem [mostly what you have below, also ADD: lack of approaches/metrics to evaluate in a more rigrious manner]. Briefly explain our current approach: (1) }

% For randomness in software, presetting the random seeds is suggested to mitigate the non-determinism. However, this approach has several drawbacks: (1) it is not trivial to select the appropriate seeds and no existing techniques treat random seeds as hyperparameters for tuning. (2) Fixing the seeds beforehand might cause the training process to converge to local optimal. Sometimes, different seeds might lead to significant differences in the results~\cite{HendersonAAAI18}. (3) Manual efforts are needed to instrument the code base for setting seeds for all any software packages with independent random number generator (e.g., numpy) and additional maintenance efforts are needed.

To address the above challenges, in this paper, we have proposed a systematic approach towards training reproducible DL models. Our approach includes a set of rigorously defined evaluation criteria, a record-and-replay-based technique for mitigating randomness in software, and a profile-and-patch-based technique for mitigating non-determinism from hardware. We have also provided a systematic guideline for DL model reproducibility based on our experience on applying our approach across different DL models. Case studies on six popular open source and one commercial DL models show that our approach can successfully reproduce the studied DL models (i.e., the trained models achieve the exact same results under the evaluation criteria). To facilitate reproducibility of our study, we provide a replication package~\cite{replicationpackage}, which consists of the implementation of open source DL models, our tool, and the experimental results. In summary, our paper makes the following contributions:

\begin{itemize}[leftmargin=*]
	\item Although there are previous research work which aimed at reproducible DL models (e.g.,~\cite{PhamASE20,LiuTOSEM20}), to the authors' knowledge, our approach is the first systematic approach which can achieve reproducible DL models during the training process. Case study results show that all the studied DL models can be successfully reproduced by leveraging our approach. 
	
	\item Compared to existing practices for controlling randomness in the software (a.k.a., presetting random seeds~\cite{PhamASE20,determined-repro}), our record-and-replay-based technique is non-intrusive and incurs minimal disruption on the existing DL development process. %In addition, the amount of overhead caused by our approach is minimal  \jack{(e.g., less than 0.5\% time and space overhead for all the studied models? \st{the time overhead and space overhead on ResNet-38 is 0.1\% and 0.2\%, respectively ).}}).
	
%	\item Our approach is also generic, as it can support DL model reproducibility for different types of DL models and DL frameworks with minimal or no modifications at all. Case study results show that the amount of overhead caused by our approach is minimal (for example, the time overhead and space overhead on ResNet-38 is 0.1\% and 0.2\%, respectively ).

	%\jack{our approach is generic, light-weight, and provide minimum discruption on the current ML dev process. It and works across various models and frameworks. Case studies on XXX shows YYY, compared to before ZZZZ}
	%Compared with existing techniques (add cites), our approach has two advantages: (1) 
	%Our process has three benefits. First, it is generic as it can be applied across different types of DL models and DL frameworks. Case studies show all the models can be successfully reproduced. Second, it is lightweight, as our results show that time overhead and space overhead are both negligible. Last, our process incurs minimal disruption on the existing DL development process, which are preferred by industrial practitioners compared to other techniques (e.g., presetting random seeds).
	% We propose a novel record and replay technique to control the randomness (i.e., the software factor) involved in DL experiments without code instrumentation. Practitioners can leverage our technique for controlling randomness if setting random seeds beforehand is not preferable. In addition, We provide a profiling method to diagnose the non-deterministic framework level function call and system level function call. 
	\item %We conduct a thorough evaluation on both open source and commercial DL models in two different domains (computer vision and time series forecasting). 
	Compared to the previous approach on verifying model reproducibility~\cite{PhamASE20}, our proposed evaluation criteria has two advantages: (1) it is more general, as it covers multiple domains (Classification and Regression tasks), and (2) it is more rigorous, as it evaluates multiple criteria, which includes not only the evaluation results on the testing dataset, but also the consistency of the training process.% and the individual prediction results. 
	
%	\item \jack{Remove if tight in space}We have conducted our case studies on DL models from two different application domains (computer vision and time series forecasting) in an industrial setting. The practitioner feedback and the lessons learned during this process would be valuable for researchers and practitioners who are also interested in this area.
	
\end{itemize}

\subsubsection*{Paper Organization}
%\boyuan{done}\jack{make sure you update this later}
%The remainder of this paper is organized as follow: 
Section~\ref{sec:background} provides background information associated with DL model reproducibility.
Section~\ref{sec:approach} describes the details of our systematic approach to training reproducible DL models.
Section~\ref{sec:result} presents the evaluation of our approach.
Section~\ref{sec:discussion} discusses the experiences and lessons learned from applying our approach.
% Section~\ref{sec:relatedwork} describes related work.
Section~\ref{sec:guideline} presents our guideline.
Section~\ref{sec:threats} describes the threats to validity of our study and Section~\ref{sec:conclusion} concludes our paper.
\section{Background and Related Work}
\label{sec:background}

In this section, we describe the background and the related work associated with constructing reproducible DL models.

\subsection{The Need for Reproducible DL models}

%\jack{P1: you start explaining things on various concepts related to repo, replicable etc. Repo in general, and then to repo in model training? Then P2: explain say repro DL models is needed by both research and practice. Start with things on open science, etc. Then say industrial now adopt these research for their product, and then say in addition to the DL models explained in research but also for their in-house models, we all need repo as: (expand the 1st paragraph of intro for each point like training, testing, debugging, by 1 - 2 sentences).}\boyuan{done}

Several terms have been used in existing literature to discuss the concepts of reproducibility in research and practice~\cite{PhamASE20,mlreprochecklist,LiuTOSEM20,HendersonAAAI18,pineau2020improving,pytorchrandom,determined-repro,tfreprorfc}. We follow the similar definitions used in ~\cite{pineau2020improving,PhamASE20}, where a particular piece of work is considered as \emph{reproducible}, if the same data, same code, and same analysis lead to the same results or conclusions.
On the other hand, replicable research refers to that different data (from the same distribution of the original data) combined with same code and analysis result in similar results.
In this paper, we focus on the reproducibility of DL models during the training process. The same training process requires the exact same setup, which includes the same source code (including training scripts and configurations), the same training and testing data, and the same environment.

Training reproducible DL models is essential in both research and practice. On one hand, it facilitates the open science movement by enabling researchers to easily reproduce the same results. Open science movement~\cite{woelfle2011open,vicente2018open} promotes sharing research assets in a transparent way, so that the quality of research manuscripts can be checked and improved.
%Open science movement~\cite{woelfle2011open,vicente2018open} promotes sharing research assets in a transparent way, so that the quality of research manuscripts can be checked and improved. Hundreds of new AI research manuscripts are posted every day, being able to quickly validate and build on top of such research benefits scientific advancement. 
On the other hand, many companies are also integrating the cutting-edge DL research into their products. Having reproducible DL models would greatly benefit the product development process. For example, if a DL model is reproducible, the testing and debugging processes would be much easier as the problematic behavior can be consistently reproduced~\cite{testMLGoogle}. In addition, many DL-based applications now require regulatory compliance and are subject to rigorous auditing processes~\cite{EuroRegulation}. It is vital that the behavior of the DL models constructed during the auditing process closely matches with that of the released version~\cite{assessmentEurope}.

\subsection{Current State of Reproducible DL Models} 

%\boyuan{I feel the related papers are not organized by software/hardware non-determinism, they mostly mention that data/code, etc. need to be consistent. Even mentioning software, they would just say use seeds. So I suggest keep it vague using two parts, the first part states the repro crisis and guideline not sufficient, the second part state describes studies.}\jack{should we divide this into three parts: (1) guidelines (a.k.a., checklists) not sufficient, as only related to snapshots of asserts, not for experiments mgmt nor related to the tooling part, (2) software (current state and problems), (3) hardware (current state and problems)}

\subsubsection{Reproducibility crisis.} In 2018, Huston~\cite{HutsonScience18} mentioned it is very difficult to verify many claims published in research papers due to the lack of code and the sensitivity of training conditions (a.k.a., the reproducibility crisis in AI). 
Similarly, Gundersen and Kjensmo~\cite{GundersenAAAI18} surveyed 400 research papers from IJCAI and AAAI and found that only 6\% of papers provided experiment code. Similarly, in software engineering research, Liu et al.~\cite{LiuTOSEM20} surveyed 93 SE research papers which leveraged DL techniques and only 10.8\% of research discussed reproducibility related issues. Isdahl and Gundersen~\cite{isdahl2019out} surveyed 13 state of the art ML platforms and found the popular ML platforms provided by well-known companies have poor support for reproducibility, especially in terms of data. Instead of verifying and reporting the reproducibility of different research work, we focus on proposing a new approach which can construct reproducible DL models.

\subsubsection{Efforts towards improving reproducibility.} 
Various efforts have been devoted to improve the reproducibility of DL models:%: from the following perspectives:
%\begin{itemize}[leftmargin=*]

	\noindent \textbf{(E1) Controlling Randomness from software.} Liu et al.~\cite{LiuTOSEM20} found that the randomness in software could impact the reproducibility of DL models and only a few studies (e.g.,~\cite{GuICSE018,ColasArxiv18,HendersonAAAI18}) reported using preset seeds to control the randomness.  
	Similarly, Pham et al.~\cite{PhamASE20} found that by controlling randomness in software, the performance variances in trained DL models decrease significantly. Sugimura and Hartl~\cite{SugimuraArxiv18} mentioned that a random seed needs to be set as a hyperparameter prior to training for reproducibility.
	Determined.AI~\cite{determined-repro}, a company that focuses on providing services for DL model training, also supports setting seeds for reproducing DL experiments. However, none of the prior studies discussed how to properly set seeds or the performance impact of different set of seeds.
	Compared to presetting random seeds, our record-and-replay-based technique to control the randomness in the software is non-intrusive and incurs minimal disruption on the existing DL development.
	
	\noindent \textbf{(E2) Mitigating non-determinism in the hardware.}
	Pham et al.~\cite{PhamASE20} discussed using environment variables to mitigate non-determinism caused by floating point rounding error and parallel computation. Jooybar et al~\cite{JooybarASPLOS13} designed a new GPU architecture for deterministic operations. However, there has been a lack of thorough assessment of the proposed solutions. In addition, our approach mainly focuses on mitigating non-determinism on common hardware instead of proposing new hardware design.
	
	\noindent \textbf{(E3) Existing guidelines and best practices.}
	To address the reproducibility crisis mentioned above, major AI conferences such as NeurIPS, ICML, and AAAI hold reproducibility workshops and advocate researchers to independently verify the results of published research papers as reproducibility challenges.
	Various documentation frameworks for DL models~\cite{GebruDatasheet18,MitchellFAT19} or checklists~\cite{mlreprochecklist} have been proposed recently. These documentations specify the required information and artifacts (e.g., datasets, code, and experimental results) that are needed to reproduce DL models. Similarly, Ghanta et al~\cite{ghanta2018systems} investigated AI reproducibility in production, where they mentioned many factors need to be considered to achieve reproducibility such as pipeline configuration and input data. Tatman et al.~\cite{tatman2018practical} indicated that high reproducibility is achieved by managing code, data, and environment. They suggest in order to reach the highest reproducibility, the runtime environment should be provided as hosting services, containers, or VMs. Sugimura and Hartl~\cite{SugimuraArxiv18} built an end-to-end reproducible ML pipeline which focuses on data, feature, model, and software environment provenance. In our study, we mainly focus on model training with the assumption that the code, data, and environment should be consistent across repeated training processes.
	However, even with consistent assets mentioned above, it is still challenging to achieve reproducibility due to the lack of tool support and neglection of certain sources of non-determinism~\cite{LeeRML19,tatman2018practical,RaffNIPS19,sugimura2018building,isdahl2019out,ghanta2018systems,PhamASE20}.
	
%\end{itemize}
% Closest to our work is the study conducted by Pham et al.~\cite{PhamASE20}, which focuses on measuring the variance caused by implementation-level non-determinism introducing factors (a.k.a the non-determinism from hardware) instead of controlling them.

\subsection{Industrial Assessment}
\huawei is a large IT company, which provides many products and services relying on AI-based components. To ensure the quality, trustworthiness, transparency, and traceability of the products, practitioners in \huawei have been investigating approaches to training reproducible DL models. We worked closely with 20 practitioners, who are either software developers or ML scientists with Ph.D degrees. Their tasks are to prototype DL models and/or productionalize DL models. We first presented the current research and practices on verifying and achieving reproducibility in DL models. Then we conducted a two hour long semi-formal interview with these practitioners to gather their opinions on whether the existing work can help them address their DL model reproducibility issues in practice. We summarized their opinions below:

%\begin{itemize}[leftmargin=*]

	\noindent \textbf{Randomness in the Software}: Practitioners are aware that currently the most effective approach to control the randomness in the software is to set seeds prior to training.
	However, they are reluctant to adopt such practice due to the following two reasons: (1) \emph{a variety of usage context:} for example, in software testing, they would like to reserve the randomness so that more issues can be exposed. However, after the issue is identified, they find it difficult to reproduce the same issue in the next run. Setting seeds cannot meet their needs in this context. (2) \emph{Sub-optimal performance:} DL models often require fine-tuning to reach the best performance. Currently, the DL training relies on certain levels of randomness to avoid local optimums. Setting seeds may have negative impacts on the model performance. Although tools like AutoML~\cite{HutterAutoML2019} have been recently widely adopted for selecting the optimal hyperparameters, there are no existing techniques which incorporate random seeds as part of their tuning or searching processes. %on tuning the performance of DL models with the presence of setting random seeds.

	%However, they did mention that customers hope to be delivered with a DL model that they could train it in a reproducible way for legal and auditing purpose. This poses the challenge that it is desireable to train reproducible DL models without presetting the random seeds.
	\noindent \textbf{Non-determinism in the Hardware}: There are research and grey literature (e.g., technical documentations~\cite{pytorchrandom}, blog posts~\cite{nvidiadeterminism}) describing techniques to mitigate the non-determinism in hardware or proposing new hardware architecture~\cite{JooybarASPLOS13}. However, in an industrial context, adopting new hardware architecture is impractical due to the additional costs and the lack of evaluation and support. In addition, the mentioned approaches (e.g., setting environment variables) are not extensively evaluated on the effectiveness and overhead. Hence, a systematic empirical study is needed before applying such techniques in practices.
	
	\noindent \textbf{Reproducibility Guidelines}: %\jack{Can we say something like they are scattered across different places (e.g., checklists are mainly for assets, software is another place, evluation is another place, etc.) on the checklist like they agree on the importance of archiving the experimental related data. but is there anything they overlooked? and say they realized the importance of archiving the assets in addition to source code and are actively using version management tools like DVC and MLFlow or developing in-house tools. also the evaluation metrics part}	
	%Practitioners from \huawei are aware that reproducibility is import in AI based software. 
	They have already applied best practices to manage the assets (e.g., code and data) used during training processes by employing data and experiment management tools. %Even they could retrieve the same code, data, and environment used in a particular training process,
	However, they found the DL models are still not reproducible. In addition, they mentioned that existing techniques in this area does not cover all of their use cases. For example, existing evaluation criteria for DL model reproducibility works for classification tasks (e.g.,~\cite{PhamASE20}), but not for regression tasks, which are the usage contexts for many DL models within \huawei. Hence, they prefer a systematic guideline which standardizes many of these best practices across various sources and usage context so that they can promote and enforce them within their organizations.

	% First, similar to the survey results reported in ~\cite{PhamASE20}, some practitioners are not aware of the importance of reproducibility. Some may even consider that training reproducible DL models is an impossible task due to the nature of DNN. Running the training processes multiple times and selecting the best-performing model is a common practice. However, such a practice may cause legal and auditing issues, as they cannot reproduce a DL model that is training months ago even with the same datasets and training code. In addition, it draws the challenge of testing and debugging the DL-based systems as they do not behave deterministically.
	% \jack{move to guideline for the evaluation metrics part} the evaluation metrics used in prior studies are not sufficient for model reproducibility. In practice, they are interested in the prediction results that are inconsistent between trained DL models with the identical training process, as it may pose potential bugs in the DL model. \jack{explain with a brief example} Current evaluation only focuses on the overall accuracy and per-class accuracy. A set of more complete evaluation metrics is needed for such issue. \jack{can we say we need wider context other than CV?}

%\end{itemize}

Inspired by the above feedback, we believe it is worthwhile to propose a systematic approach towards training reproducible DL models. We will describe our approach in details in the next section. %Such approach could enhance DL-based systems in many aspects.
\section{Our approach}
\label{sec:approach}

Here we describe our systematic approach towards reproducing DL models. 
Section~\ref{sec:approach:overview} provides an overview of our approach. Section~\ref{sec:approach:phase1} to ~\ref{sec:approach:phase5} explain each phase in detail with a running example.

\subsection{Overview}
\label{sec:approach:overview}

There are different stages in the DL workflow~\cite{AmershiSEIP19}. The focus of our paper is training reproducible DL models. Hence, we assume the datasets and extracted features are already available and can be retrieved in a consistent manner. 
% Our approach only focuses on the model training and evaluation stage.%\boyuan{i think training process is an instance of model training, so we should be good}\jack{we keep saying the training process, would this be conflict?}

%\boyuan{done}\jack{Provides an overview of development process saying includes various steps like data collection, etc. (cite MS paper)). Then state that since the focus of this paper is on training DL models, we assume there is datasets and associated features are available etc. Our process is concentrated with the model training phase}

\begin{figure*}[h]
%\begin{figure}[h]
	\centering
	\includegraphics[scale=0.55]{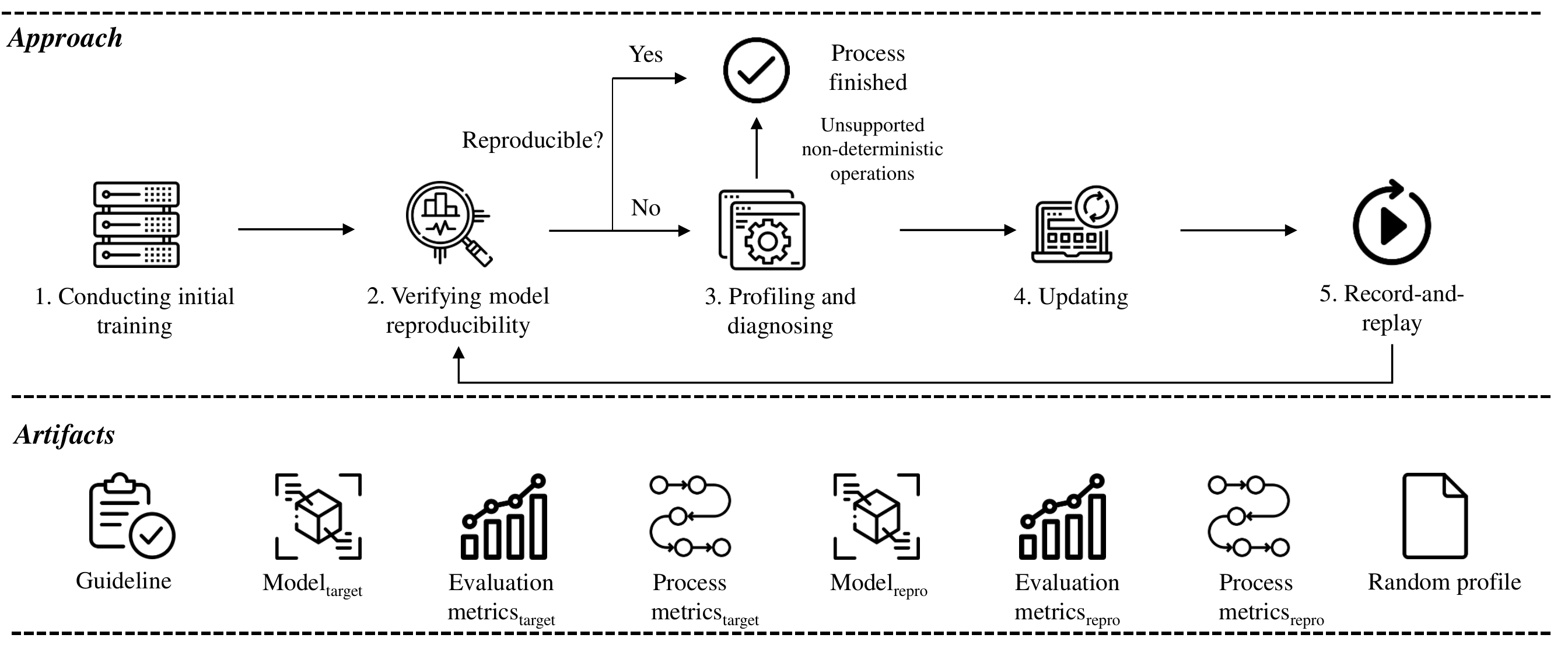}
	\caption{An overview of our approach.}
	\label{fig:controlapproach}
	\vspace{-2mm}
%\end{figure}
\end{figure*}

%\boyuan{done}\jack{just 1 sentence for each phase is okay. pls shrink the ones below.}
Figure~\ref{fig:controlapproach} presents the overview of our approach, which consists of five phases. (1) During the \emph{Conducting initial training} phase, we prepare the training environment and conduct the training process twice to generate two DL models: $Model_{_{target}}$ and $Model_{_{repro}}$. (2) During the \emph{Verifying model reproducibility} phase, the two DL models from the previous phase are evaluated on a set of criteria to check if they yield the same results. If yes, $Model_{_{target}}$ is reproducible and the process is completed. We also will update the reproducibility guideline if there are any new mitigation strategies that have been introduced during this process. If not, we will proceed to the next phase. (3) During the \emph{Profiling and diagnosing} phase, the system calls and function calls are profiled. Such data is used to diagnose and identify the root causes behind non-reproducibility. (4) During the \emph{Updating} phase, to mitigate newly identified sources of non-determinism, the system calls that need to be intercepted by the record-and-replay technique are updated and the non-deterministic operations due to hardware are patched. (5) During the \emph{Record-and-replay} phase, the system calls, which introduce randomness during training, are first recorded and then replayed. Two DL models, $Model_{_{target}}$ and $Model_{_{repro}}$, are updated with the DL models during the recording and replaying steps, respectively. %constructed during this phrase is considered as the new $Model_{_{target}}$. (6) During the \emph{Replaying} phase, we repeat the same training process, during which the recorded random values are replayed. The DL model constructed during this phrase is considered as the new $Model_{_{repo}}$. 
These two updated DL models are verified again in Phase 2. This process is repeated until we have a reproducible DL model.

% \jack{just 1 sentence for each phase is okay. pls shrink the ones below. Discussion: do we want to add a (0) Setting up Training Environment phase?} (1) During the \emph{Reproducibility Verification} phase, the DL model training process is repeated twice to verify if the DL models can be reproduced; (2) During the \emph{Profiling and Diagnosing} phase, \jack{fill in the rest}

To ease explanation, in the rest of the section, we will describe our approach using LeNet-5 as our running example. LeNet-5~\cite{LecunIEEE98} is a popular open source DL model used for image classification. The dataset used for training and evaluation is MNIST~\cite{mnist}, which consists of a set of 60,000 images for training, 10,000 images for testing. Each image is assigned a label representing the handwritten digits from 0 to 9.

% by demonstrating how the training process for LeNet-5, a popular DL model, can be reproduced. LeNet-5~\cite{LecunIEEE98} is a very popular open source image classification model which uses MNIST as its training and evaluation dataset. 

%\textbf{Running Example.} In the rest of the section, we will describe each phase in detail with a running example.
%The DL algorithm we use for the training is LeNet-5~\cite{LecunIEEE98}, a popular open source image classification algorithm. \jack{move to phase 1} The dataset used for training and evaluation is MNIST~\cite{LecunIEEE98}. We split the datasets into three parts: training, validation, and testing similar to the prior work~\cite{PhamASE20}. There are 60,000 instances in the training datasets, 7,500 in the validation datasets, and 2,500 in the testing datasets. 
%To perform the training, we use a SUSE Linux Enterprise Server 12 machine with a Tesla-P100-16GB GPU. The GPU related libraries are CUDA 10.0.130 and CuDNN 7.5.1. The DL framework we use is Tensorflow 1.14 and Keras 2.24. We use Conda environment with Python 3.6 to ensure the software dependencies are consistent.
 
\subsection{Phase 1 - Conducting initial training}
\label{sec:approach:phase1}

The objective of this phase is to train two DL models under the same experimental setup. %set up a training environment and make sure the assets used in training \jack{and evaluation?} are consistent \jack{across multiple runs}. 
This phase can be further broken down into the following three steps: %(1) setting up experimental assets; and (2) checking the consistency of experimental assets.

%\jack{within phase 0 (conducting initial training experiments), we will have these steps: (1) environment setup (also touch on data splitting), (2) training target DL model, (3) verifying assets and re-training}

%\begin{smallitemize}
\vspace{-1mm}
	\subsubsection*{Step 1 - Setting up the experimental environment} 
%\noindent\textit{Step 1 - Setting up the experimental environment.}
	In this step, we set up the experimental environment, which includes downloading and configuring the following experimental assets: the dataset(s), the source code for the DL model, and the runtime environment based on the required software dependencies and the hardware specifications~\cite{IdowuSEIP21}. Generally the experimental assets are recorded in documentations like research papers, reproducibility checklist~\cite{mlreprochecklist}, or model cards~\cite{MitchellFAT19} and data sheets~\cite{GebruDatasheet18}. For our running example, documentations are from research papers~\cite{LecunIEEE98,PhamASE20}. The code for LeNet-5 is adapted from a popular open source repository~\cite{bigballon2017cifar10cnn}, and the MNIST dataset is downloaded from~\cite{mnist}. We further split the dataset into three parts: training, validation, and testing similar to the prior work~\cite{PhamASE20}. In particular, we split the 10,000 images in testing into 7,500 images and 2,500 images. The 7,500 images are used for validation in the training process, and the 2,500 images are used to evaluate the final model, which are not exposed to the training process.
	We deploy the following runtime environment: for the software dependencies, we use Python 3.6 with TensorFlow 1.14 GPU version. For the hardware specification, we use a SUSE Linux Enterprise Server 12 machine with a Tesla-P100-16GB GPU. The GPU related libraries are CUDA 10.0.130 and CuDNN 7.5.1.

%	\item \emph{Step 2 - training the target DL model}: 
\vspace{-1mm}
\subsubsection*{Step 2 - Training the target DL model}
%\noindent\textit{Step 2 - Training the target DL model.}
	In this step, we invoke the training scripts to generate the target DL model, called $Model_{_{target}}$. During the training process, we collect the following set of metrics: loss values, the training epochs, and the training time. This set of metrics is called $ProcessMetrics_{_{target}}$. In our running example, we invoke the training scripts for LeNet-5 to construct the DL model and record its metrics. 

%	\item \emph{Step 3 - verifying assets and retraining}:
\vspace{-1mm}
\subsubsection*{Step 3 - Verifying assets and retraining}
%\noindent\textit{Step 3 - Verifying assets and retraining.}
	In this step, we first verify whether the experimental assets are consistent with the information provided in step 1. There are many approaches to verifying the experimental assets. For example, to verify the dataset(s), we check if the SHA-1 checksum is consistent. To verify the software environment, we check the software dependency versions by reusing the same environment (e.g., docker, VM) or simply checking all the installed software packages by commands like \texttt{pip list}. Once the assets are verified, we perform the same training process as step 2 to generate another DL model, named as $Model_{_{repro}}$. We also record the same set of metrics, called as $ProcessMetrics_{_{repro}}$, during the training process. The two DL models along with the recorded set of metrics will be used in the next phase for verifying model reproducibility. In our running example, we reuse the same experimental environment without modifying the source code and the datasets to ensure the asset consistency. Then we repeat the training process to collect the second LeNet-5 model and its metrics.

%\end{smallitemize}

\subsection{Phase 2 - Verifying model reproducibility}
\label{sec:approach:phase2}
%\jack{for each phase, you need to (1) explain the high level objective of this phase, explain any assumptions for this phase, and the types of input and output (if applicable) (2) then say it's further broken down into x steps. (3) explain each steps in details with running examples }\boyuan{done}

The objective of this phase is to verify if the current training process is reproducible by comparing the two DL models against a set of evaluation criteria. %could yield reproducible DL models. The inputs of this phase are the experimental assets checked in the previous phase. The outputs of this phase are two DL models and the corresponding evaluation metrics. 
This phase consists of the following three steps: % (1) training DL models twice and collecting the runtime data; (2) summarize the evaluation metrics; and (3) report if the DL models are reproducible. 

%\jack{phase 1, verifying model reproduciability (1) evaluation, (2) report results}

%\begin{smallitemize}
%	\item \emph{Step 1 - Verifying the reproduciability of the training results}: 
%\subsubsection*{Step 1 - Verifying the reproducibility of the training results}:
\vspace{-1mm} 
\subsubsection*{Step 1 - Verifying the reproducibility of the training results} 
	In this step, we evaluate the two DL models, $Model_{_{target}}$ and $Model_{_{repro}}$ on the same testing dataset. Depending on the tasks, we use different evaluation metrics: 
	
	\begin{itemize}[leftmargin=*]
		\item \emph{Classification tasks}: For classification tasks, we evaluate three metrics: the overall accuracy, per-class accuracy, and the prediction results on the testing dataset. Consider the total number of instances in testing datasets is $N_{_{test}}$. The number of correctly labeled instances is $N_{_{correct}}$. For label $i$, the number of instances are $N_{_{test_{i}}}$. The correctly labeled instances of label $i$ is $N_{_{correct_{i}}}$. Hence, the overall accuracy is calculated as:
		%\begin{equation}
		$\textit{Overall accuracy} = \frac{N_{_{correct}}}{N_{_{test}}}$.
		For each label $i$, the per-class accuracy is calculated as:
		%\begin{equation}
		$\textit{Per-class accuracy (label i)} = \frac{N_{_{correct_{i}}}}{N_{_{test_{i}}}}$.
		%\end{equation} 
		In addition, we collect the prediction results for every instance in the testing dataset. %\boyuan{I feel no need to use  }\jack{How do you call them, P\_i?}
				
		\item \emph{Regression tasks}: For regression tasks, we evaluate the Mean Absolute Error (MAE). The total number of instances in the testing dataset is $N_{test}$ Consider for each instance, the true observed value is $X_{i}$ and the predicted value is $Y_{i}$. MAE is calculated as:
		%\begin{equation}
		$MAE =  \frac{\sum_{i=1}^{N_{test}} |Y_{i} - X_{i}|}{N_{test}}$.
		%\end{equation}
		These metrics are called as $EvaluationMetrics_{_{target}}$ and $EvaluationMetrics_{_{repro}}$ for these two models, respectively. In our running example, we use evaluation metrics for classification tasks as LeNet-5 is used for image classification.
		
	\end{itemize}
	
%	\item \emph{Step 2 - Verifying the reproduciability of the training process}:
\vspace{-1mm} 
\subsubsection*{Step 2 - Verifying the reproducibility of the training process}
	In this step, we compare the collected metrics for $Model_{_{target}}$ and $Model_{_{repro}}$ (i.e., $EvaluationMetrics_{_{target}}$ vs. $EvaluationMetrics_{_{repro}}$ and $ProcessMetrics_{_{repro}}$ vs. $ProcessMetrics_{_{target}}$) by a Python script.
	For evaluation metrics, we check if $EvaluationMetrics_{_{target}}$ and $EvaluationMetrics_{_{repro}}$ are exactly identical. For process metrics, we check if the loss values during each epoch, and the number of epochs are the same.
	
%	\item \emph{Step 3 - Reporting the results}: 
\vspace{-1mm}
\subsubsection*{Step 3 - Reporting the results}
	A DL model is reproducible if both the evaluation metrics and the process metrics are identical (except for the training time). If the DL models are not reproducible, we move on to the next phase.
	
%\end{smallitemize}

%In the first step, we start the training process twice by simply invoking the same training scripts twice. During this step, we collect the loss values during training, the training epochs, the training time, and the evaluation metrics. In the second step, we compare the two sets of evaluation metrics from two DL models. The evaluation metrics for classification tasks and regression tasks are defined below.

%In our running example, we start the training process by invoking the training scripts via command \texttt{python3 mnist\_lenet\_5.py}. The DL model LeNet-5 is conducting a classification task on the images of handwritten digits. Hence, we evaluate the overall accuracy, per-class accuracy, and the prediction results. 

In our running example, the two DL models emit different evaluation metrics. The overall accuracy for the two models are 99.16\% and 98.64\%, respectively.
For per-class accuracy, the maximum absolute differences could be as large as 2.3\%.
Among the 2,500 prediction results, 48 of them are inconsistent.
None of the loss values during the epochs are the same. The total number of training epochs are 50 as it is pre-configured. 
This result shows that the two DL models are not reproducible. Hence, we proceed to the next phase.

\subsection{Phase 3 - Profiling and diagnosing}
\label{sec:approach:phase3}
The objective of this phase is to identify the rationales on why the DL models are not reproducible through analysis of the profiled results. %We will repeat the training process with profiling techniques and diagnose the rationale based on the profiling results. 
The output of this phase is a list of system calls that introduce software-related randomness and a list of library calls that introduce hardware related non-determinism. This phase consists of the following four steps: 
%(1) system call profiling; (2) diagnosing sources of randomness; (3) library call profiling; (4) diagnosing sources of non-determinism in hardware.

\vspace{-1mm}
\subsubsection*{Step 1 - Profiling}

This step is further divided into two sub-steps based on the type of data, which is profiled: 

%\begin{itemize}[leftmargin=*]
	\noindent\emph{Step 1.1 - Profiling system calls}: After inspecting the documentation and the source code of the DL frameworks, we have found that the randomness from software can be traced to the underlying system calls. 
	For example, in TensorFlow, the random number generator is controlled by a special file (e.g., \texttt{/dev/urandom}) in the Linux environment.
	When a random number is needed in the training, the kernel will invoke a system call to query \texttt{/dev/urandom} for a sequence of random bytes. The sequence of random bytes is then used by the random generation algorithm (e.g., the Philox algorithm~\cite{SalmonSC11}) to generate the actual random number used in the training process. 
	
	\noindent \emph{Step 1.2 - Profiling library calls}: To mitigate the sources of non-determinism in the hardware, popular DL frameworks start to provide environment variables to enhance reproducibility. For example, in TensorFlow 2.1 and above, setting the environment variable \texttt{TF\_CUDNN\_DETERMINISTIC} to be "true" could indicate the cuDNN libraries to disable the auto-tuning feature and use the deterministic operations instead of non-deterministic ones. 
	However, there are still many functions that could introduce non-determinism even after the environment variable is set. In addition, lower versions of TensorFlow (e.g., 1.14), which does not support such configuration, are still widely used in practice. To address this issue, Nvidia has released an open source repository~\cite{nvidiadeterminism} to document the root causes of the non-deterministic functions and is currently working on providing patches for various versions of TensorFlow. Not all the operations could be made deterministic and ongoing efforts are being made~\cite{tfreprorfc}.
	Hence, to diagnose the sources of non-determinism in hardware, we perform function level profiling to check if any of the functions are deemed as non-deterministic. Different from profiling the system calls, which extracts call information at the kernel level, the goal of profiling the library calls is to extract all the invoked function calls at framework level (e.g., \texttt{tensorflow.shape}).
	
%\end{itemize}

In our running example, we repeat the training process of LeNet-5 with the profiling tools. We use \texttt{strace} to profile the list of system calls invoked during the training process. \texttt{strace} exposes the interactions between processes and the system libraries and lists all the invoked system calls. 
We use \texttt{cProfile}, a C-based profiling tool, to gather the list of invoked functions at the framework level. 

\vspace{-1mm}
\subsubsection*{Step 2 - Diagnosing sources of randomness}
\begin{figure}[]
	\centering
	\includegraphics[scale=0.42]{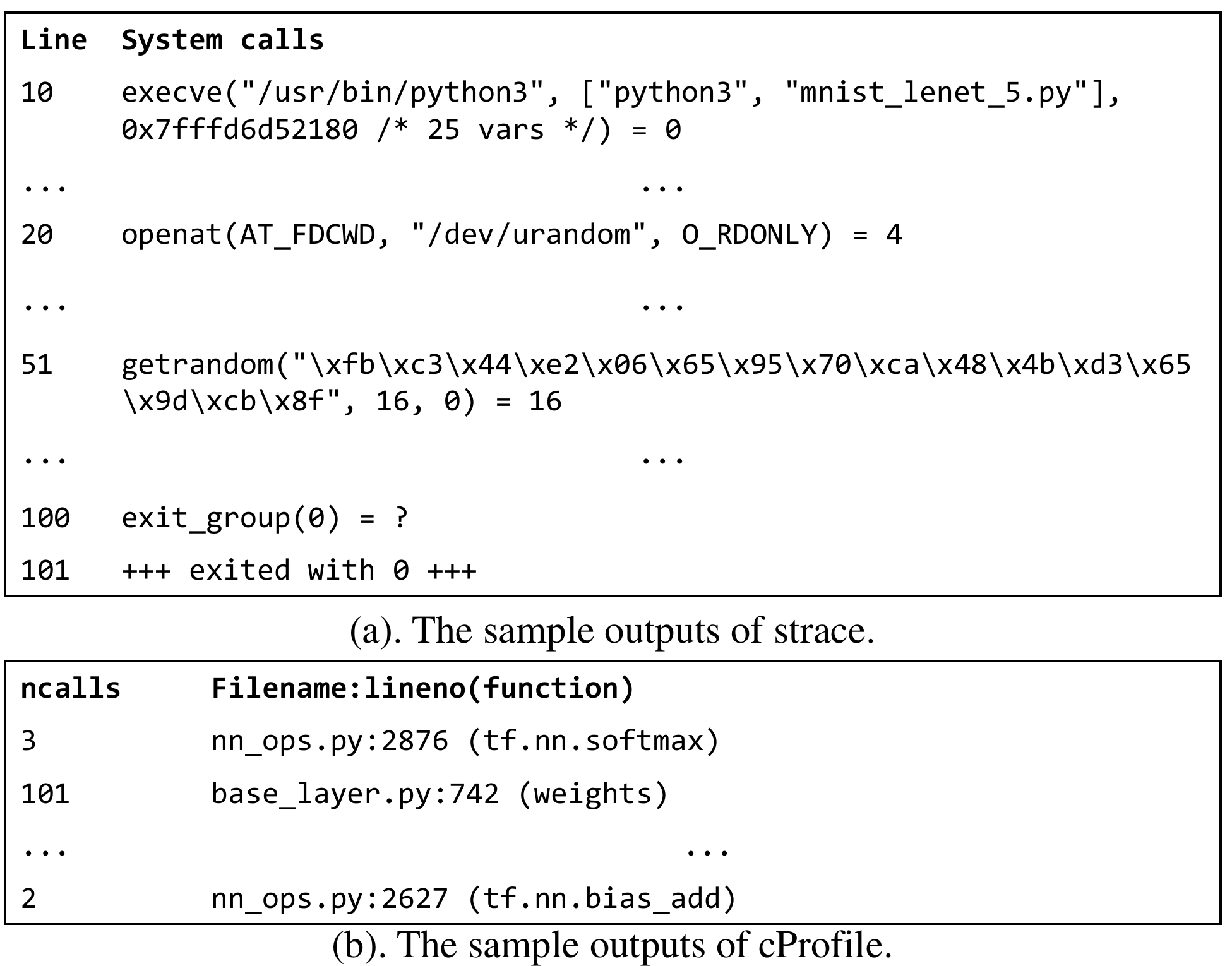}
		
	\caption{Sample output snippets from \texttt{strace} and \texttt{cProfile}.}

	\label{fig:strace_cprofile}
	\vspace{-5mm}
\end{figure}

In this step, we analyze the recorded data from \texttt{strace} to identify the set of system calls which can contribute to software-related randomness. We consult with the documentation of system calls and identify the list of system calls, which causes randomness. This list varies depending on the versions of the operating systems. For example, the system call \texttt{getrandom} is only used in later version of Linux kernel (version 3.17 and after). Prior to 3.17, only \texttt{/dev/urandom} is used. Hence, we have to not only search for the list of randomness introducing system calls in the \texttt{strace} data, but also checking if the function parameters contain \texttt{"/dev/urandom"}. 
Figure~\ref{fig:strace_cprofile}(a) shows a snippet of the sample outputs from \texttt{strace} in our running example. Each line corresponds to one system call. For example, line 10 shows that the program from \texttt{/usr/bin/python3} is executed with the script \texttt{mnist\_lenet\_5.py} and the return value is 0. 
The system call recorded at line 20 reads from \texttt{"/dev/urandom"}, and system call (\texttt{getrandom}) recorded at line 51 is also invoked. Both of the two system calls introduce software-related randomness. %By reading the official documentation from the system library in Linux (i.e., GlibC), we confirm that \texttt{getrandom} also returns random bytes. In addition, \texttt{getrandom} is only introduced in version 3.17 of Linux kernel. Prior to 3.17, only \texttt{/dev/urandom} is used. This also highlights the importance of consistency of runtime environment. The last system call at line 100 marks the end of this process.

\vspace{-1mm}
\subsubsection*{Step 3 - Diagnosing sources of non-determinism in hardware}
In this step, we cross-check with the Nvidia documentation~\cite{nvidiadeterminism} to see if any of the library functions invoked during the training process triggers the non-determinism functions at the harware layer. If such functions exist, we check if there is a corresponding patch provided. If no such patch exists, we will document the unsupported non-deterministic operations and finish the current process. If the patch exists, we will move on to the next phase. Figure~\ref{fig:strace_cprofile}(b) shows a snippet of the sample outputs of \texttt{cProfile} for our running example.
%We omit the several fields in the original cProfile outputs as they are not relevant to our study. In the outputs of cProfile, we can inspect the function calls invoked by the DL framework.
The functions \texttt{softmax}, \texttt{weights}, \texttt{bias\_add} are invoked 3, 101, and 2 times, respectively. We find that \texttt{bias\_add} leverages the CUDA implementation of \texttt{atomicAdd()}, which is commonly used in matrix operations. The behavior of \texttt{atomicAdd()} is non-deterministic because of the order of parallel computations is undetermined, which causes rounding error in floating point calculation~\cite{nvidiadeterminism,PhamASE20}. The other function calls do not trigger non-deterministic behavior.%\jack{discuss the unsupport stuff briefly?}

%\begin{figure}[h]
%	\centering
%	\includegraphics[scale=0.55]{pics/cprofile.pdf}
%	\caption{The sample outputs of cProfile}
%	\label{fig:cprofile}
%\end{figure}

\subsection{Phase 4 - Updating}
\label{sec:approach:phase4}
In this phase, we update our mitigation strategies based on the diagnosis results from the previous phase. % The inputs of this phase is the identified list of functions that introduce randomness and the list of functions that are non-deterministic due to hardware. Updated mitigation strategies for the two aspects will be outputted after this phase. In particular, 
This phase can be further broken down into two steps: %(1) updating the recorded system calls; and (2) applying patches for non-deterministic library calls. 
\vspace{-1mm}
\subsubsection*{Step 1 - Updating the list of system calls for recording}
For the randomness introducing functions, we will add them into the list of intercepted system calls for our record-and-replay technique, so that the return values of the relevant system calls can be successfully recorded (described in the next phase). In our running example, we will add the invocation of reading \texttt{/dev/urandom} and \texttt{getrandom} into the list of intercepted system calls to mitigate randomness in the software. 

\vspace{-1mm}
\subsubsection*{Step 2 - Applying the right patches for non-deterministic library calls.}
For the non-deterministic functions related to hardware, we check if there are existing patches that address such problems and integrate them into the training scripts. In our running example, after checking the documentation from the Nvidia repository~\cite{nvidiadeterminism}, we found one patch, which replaces \texttt{bias\_add} calls with \texttt{\_patch\_bias\_add}. We then integrated the patch to the source code of the training scripts by adding these two lines of code: \texttt{from tfdeterminism import patch} and \texttt{patch()}. In this way, during the subsequent training process of LeNet-5, the non-deterministic functions will be replaced with the deterministic alternatives. 

\subsection{Phase 5 - Record-and-Replay}
\label{sec:approach:phase5}
\begin{figure}[]
	\centering
	\includegraphics[scale=0.55]{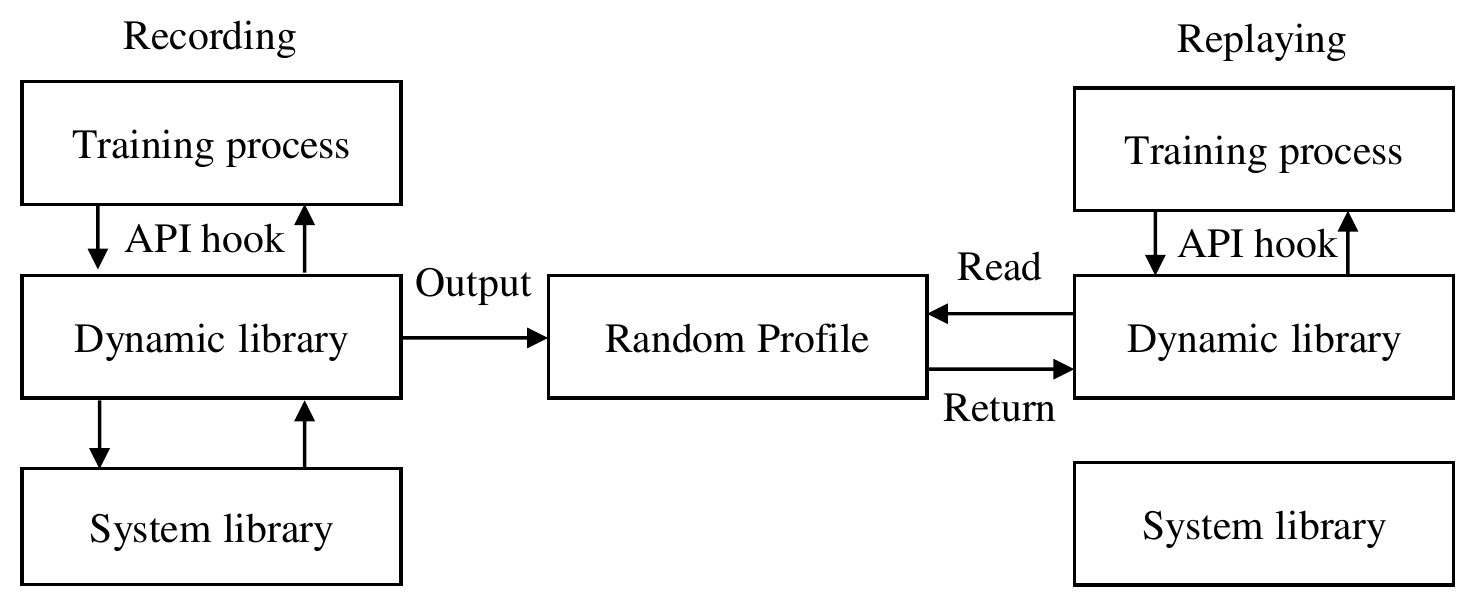}
	\caption{Our record-and-replay technique.}
	\label{fig:record}
	\vspace{-3mm}
\end{figure}

As explained in Section~\ref{sec:background}, presetting random seeds is not preferred by practitioners due to various drawbacks. There are libraries (e.g., numpy) which support the recording and replaying of random states through explicit API calls. 
%Figure~\ref{fig:recordstate} shows such an example for \texttt{numpy} library. By leveraging the API pair \texttt{getstate} and \texttt{setstate}, the same sequences of random numbers can be generated. 
However, this method is also intrusive and would incur additional costs we described before. More importantly, mainstream DL frameworks such as TensorFlow and PyTorch do not provide such functionality. 
Hence, we propose a record-and-replay technique (overview shown in Figure~\ref{fig:record}) to address these challenges. This phase has two steps: 

\vspace{-1mm}
\subsubsection*{Step 1 - Recording}
In this step, we record the random values returned by system calls during the training process. We will run the identical training process as in Phase 1 with the our recording technique enabled. We leverage the API hook mechanism to intercept the system calls by pointing the environment variable \texttt{LD\_PRELOAD} to our self-implemented dynamic library. It tells the dynamic loaders to look up symbols in the dynamic library defined in \texttt{LD\_PRELOAD} first. The functions of the dynamic library will be first loaded into the address space of the process. 
Our dynamic library implements a list of functions which have the same symbols of the randomness introducing system calls in the system libraries. These self-implemented functions will be loaded first and invoke the actual randomness introducing system calls to get the returned random bytes.
The sequences of random bytes emitted by the system calls are then recorded into an user-defined object. These objects are then serialized and written into files called the random profile. We replace $Model_{_{target}}$ and $ProcessMetrics_{_{target}}$ with the DL model and the process metrics generated in this step. 
%We also collect all the runtime data \boyuan{changed}\jack{and statistics during the training process?} as mentioned in Phase 1. In addition, we keep the execution logs to monitor if the randomness is being successfully recorded.

In our running example, two types of system calls are intercepted (i.e., \texttt{getrandom} and the read of \texttt{/dev/urandom}) and the return values are successfully recorded. %Figure~\ref{fig:logs}(a) shows a snippet of the execution logs that are generated from dynamic library.
The outputted random profile is stored at a pre-defined path in the local file system called \texttt{urandom.conf} and \texttt{getrandom.conf}. For the process-related metrics, we collect the loss values for each epoch (e.g., the loss value of the first epoch is 1.062), the training time (106.9 seconds), and the number of training epochs (50).

%\begin{figure}[h]
%	\centering
%	\includegraphics[scale=0.40]{pics/logs.pdf}
%	\caption{The execution logs collected in the record-and-replay phase.}
%	\label{fig:logs}
%\end{figure}

%\subsection{Phase 5 - Replay}
%\label{sec:approach:phase5}
\vspace{-1mm}
\subsubsection*{Step 2 - Replaying}
In this step, we repeat the same training process as the previous step while replaying the random values stored in the random profile by leveraging the API hook mechanism.  
%while enabling the API hook mechanism. The objective of this phase is to replay the random values stored in the random profile in the training process. The output of this phase is a trained DL model $Model_{replay}$. We will also collect the runtime data similar to the recording phase. 
As shown in Figure~\ref{fig:record}, our dynamic library will search for existing random profile. If such random profile exists, the recorded random bytes are used to to replace the random bytes returned by the system calls. We also replace $Model_{_{repro}}$ and $ProcessMetrics_{_{repro}}$ with the DL model and the process metrics generated in this step. In our running example, we compare the execution logs between our recording and replaying steps and verify that the same set of random numbers are generated in these two steps.
%as highlighted with red dashed boxes in both sub-figures of Figure~\ref{fig:logs}, exactly

Once this phase is completed, the two updated DL models are sent to Phase 2 for verifying their reproducibility again. This process is repeated until the DL model is shown to be reproducible or we find certain sources of non-determinism that currently do not have existing solutions. For example, the function \texttt{tf.sparse.sparse\_den\_m\\atmul} is noted in ~\cite{nvidiadeterminism} that no solution has been released yet. The reasons for non-reproducibility should be included in the documentations along with released DL models.%\boyuan{done}\jack{should we have some output after diagnosis? how do we deal with this?} 

In our running example, in terms of $Model{_{repro}}$ and $Model{_{target}}$, $EvaluationMetrics_{_{repro}}$ and $EvaluationMetrics_{_{target}}$ are identical (i.e., overall accuracy, the per-class accuracy, and prediction results on the testing datasets of two DL models are identical). Except for the training time, the $ProcessMetrics_{_{repro}}$ and $ProcessMetrics_{_{target}}$ are also identical. In conclusion, we consider the trained LeNet-5 models to be reproducible.

\section{Results}
\label{sec:result}
In this section, we evaluate our approach against open source and commercial DL models. 
Section~\ref{sec:casestudy} describes our case study setup. Section~\ref{sec:evaluate} presents the analysis of our evaluation results.

\subsection{Case Study Setup}
\label{sec:casestudy}

We have selected six commonly studied Computer Vision (CV) related DL models similar to prior studies~\cite{PhamASE20,MaDeepGaugeASE18,MaFSE18,GuoASE19,GuIFIP21}. The implementations of these models are adapted from a popular open source repository used by prior studies~\cite{bigballon2017cifar10cnn,GuoASE19,GuIFIP21,MaFSE18}. %They have been carefully examined by the authors of this paper to make sure there are no implementation bugs.

Table~\ref{tab:pjsetup} shows the details about the studied datasets and the models. The studied models are LeNet-1, LeNet-4, LeNet-5, ResNet-38, ResNet-56, and WRN-28-10.
These models mainly leverage the Convolutional Neural Network (CNN) as their neural network architectures. We use popular open source datasets like MNIST, CIFAR-10, and CIFAR-100. The models and datasets have been widely studied and evaluated in prior SE research~\cite{MaFSE18,GerasimouICSE20,GuoASE19,MaDeepGaugeASE18}. %, as a prior study~\cite{PhamASE20} shows that 64\% of surveyed SE papers use at least one of the models.%maybe delete this sentence. 
For models in LeNet family, we train for 50 epochs. For models in ResNet family and WRN-28-10, we train for 200 epochs~\cite{PhamASE20}.

We also study ModelX used in a commercial system from \huawei. ModelX is a LSTM-based DL model used to forecast energy usages. ModelX is trained with the early-stopping mechanism and the epochs are not deterministic. The training process will automatically stop when the loss values have not improved for 5 epochs. The maximum number of epochs in the training is set to be 50.
ModelX uses proprietary time-series data as their training and testing datasets and is deployed in systems, which are used by tens of millions of customers.
Due to the company policy and review standards, we cannot disclose the detail design of the DL model. The implementation of other open source models is disclosed in our replication package~\cite{replicationpackage}.

\begin{table}[]
	\centering
	\caption{The DL models used in our case study.}
	
	% \usepackage{multirow}

		\resizebox{\columnwidth}{!}{\begin{tabular}{l|l|c|c|c}
			\toprule
			\textbf{Models}    & \textbf{Datasets}                  & \textbf{\# of  Labels}       & \textbf{Setup}                            & \textbf{Task}          \\
			\midrule
			LeNet-1~\cite{LecunIEEE98}   & \multirow{3}{*}{MNIST~\cite{mnist}}    & \multirow{3}{*}{10} & \multirow{5}{*}{All} & \multirow{6}{*}{Classification}  \\
			LeNet-4~\cite{LecunIEEE98}   &                           &                     &                                 &                      \\
			LeNet-5~\cite{LecunIEEE98}   &                           &                     &                                 &                      \\
			\cline{1-3}
			ResNet-38~\cite{HeCVPR16} & \multirow{2}{*}{CIFAR-10~\cite{cifar}} & \multirow{2}{*}{10} &                                 &                      \\
			ResNet-56~\cite{HeCVPR16} &                           &                     &                                 &                      \\
			\cline{1-4}
			
			WRN-28-10~\cite{WRN16} & CIFAR-100~\cite{cifar}                 & 100                 &                G1-G10                 &                \\
			\hline
			ModelX   & Dataset X                 & -                   &  G1-G5                      &  Regression             \\
			\bottomrule
		\end{tabular}}

	\label{tab:pjsetup}
\end{table}

For both open source and commercial models, we perform the training processes with different setups. In total, we have 16 different setups listed in Table~\ref{tab:expsetup}: 

\begin{itemize}[leftmargin=*]
	\item First, there are two general groups of setups, CPU-based and GPU-based, to assess whether our approach can address different sources of hardware-related non-determinism.
	For CPU-based experiments (i.e., C1 - C6), we only train the models of the LeNet family and ResNet family, as the training for WRN-28-10 and the commercial project takes extremely long time (longer than a week) and not practical to use in field. 
	For GPU-based experiments (i.e., G1 - G10), we conduct experiments on training all the aforementioned models. 
	The CPU used for the experiments is Intel(R) Xeon(R) Gold 6278C CPU with 16 cores and the GPU we use is Tesla-P100-16GB.
	The GPU related libraries are (CUDA 10.0.130 and cuDNN 7.5.1), and (CUDA 10.1 and cuDNN 7.6) for TensorFlow 1.14 and TensorFlow 2.1, respectively.
	We use two sets of hardware related libraries due to compatibility issues mentioned in the official TensorFlow documentation~\cite{tfversion}.
	
	\item Then, within the same hardware setup, we also conduct experiments by varying the software versions.
	For open source models, we use both TensorFlow 1.14 and TensorFlow 2.1. We choose to carry out our experiments on these two TensorFlow versions as major changes~\cite{tfmigrate} have been made from TensorFlow 1.X to TensorFlow 2.X and there are still many models which use either or both versions. Hence, we want to verify if our approach can work with both versions. 
	For ModelX, we only use TensorFlow 1.14 as it currently only supports the TensorFlow 1.X APIs. % In addition, ModelX has an early stopping mechanism, which stops the training process if the loss values have not been improved for 5 epochs.
	
	\item For CPU-based experiments in a particular software version (e.g., TensorFlow 1.14), we have three setups: C1 is to run the training process without setting seeds and without enabling the record-and-replay technique. C2 is to run the training with seeds, whereas C3 is to run the training with record-and replay enabled.
	
	For GPU-based experiments in a particular software version (e.g., TensorFlow 1.14), we have five setups: G1 and G2 are similar to C1 and C2. G3 is to run the experiments with patching only to evaluate the variance related to software randomness. G4 and G5 are both running with patches, but configured with either setting seeds or enabling record-and-replay, respectively.
\end{itemize}

\begin{table}[]
	\centering
	\caption{The information of all the experiment setups. R\&R represents Record-and-Replay.}
	\begin{tabular}{lccccc}
		\toprule
		\textbf{ID}  & \textbf{Hardware} & \textbf{Software} & \textbf{Seed} & \textbf{R \& R} & \textbf{Patch} \\
		\midrule
		C1  & \multirow{6}{*}{CPU}       & \multirow{3}{*}{TF1.14}   & -              & -                   & -            \\
		C2  &       &    &   Yes           & -                   & -            \\
		C3  &       &    & -              & Yes                  & -            \\
		\cline{3-6}
		C4  &       & \multirow{3}{*}{TF2.1}    & -              & -                   & -            \\
		C5  &       &     & Yes              & -                   & -            \\
		C6  &       &     & -              & Yes                   & -            \\
		\hline
		G1  & \multirow{10}{*}{GPU}      & \multirow{5}{*}{TF1.14}   & -              & -                   & -            \\
		G2  &       &    & Yes             & -                   & -            \\
		G3  &       &    & -              & -                  & Yes            \\
		G4  &       &    & Yes             & -                   & Yes           \\
		G5  &       &    & -              & Yes                  & Yes           \\
		\cline{3-6}
		G6  &       & \multirow{5}{*}{TF2.1}    & -              & -                   & -            \\
		G7  &       &     & Yes             & -                   & -            \\
		G8  &       &     & -              & -                  & Yes            \\
		G9  &       &     & Yes             & -                   & Yes           \\
		G10 &       &     & -              & Yes                  & Yes           \\
		\bottomrule
	\end{tabular}
	\label{tab:expsetup}

\end{table}

For each setup, we run the experiments 16 times similar to a prior study~\cite{PhamASE20}. 
The training dataset is split into batches and fed into the training process; the validation dataset is used for evaluating the losses during training; and the testing dataset is used for evaluating the final models. In other words, the training and validation dataset are known to the trained DL models, while the testing data is completely new to the model to mimic the realistic field assessment.

We further divide the 16 runs of DL experiments into 8 pairs, each of which consists of two runs. We then compare the evaluation metrics from each pair of runs to verify reproducibility. For the setups with random seeds configured, we choose 8 most commonly used random seeds for each pair (e.g., 0 and 42)~\cite{commonseeds}. We collect the process and evaluation metrics as described in Section~\ref{sec:approach:phase2}.
% The first run is the original training process and the second run is the validation training process\jack{do you mean trained DL models through our approach?}.
% Our goal is to evaluate if the models generated from the original training process is identical to those generated from the validation training process based on our verification criteria. 

% For example, in G2, where we will set the random seeds, we set the same random seed in the same group (e.g., 0) and compare the generated models. In the setups where we will use the record and replay mechanism, we will group the record run and the replay run in the same group to compare the evaluation metrics. \jack{add the evaluation metrics (e.g. the repo checking metrics, the overhead related metrics, etc.)}

% For the six open source DL algorithms, we collect the overall accuracy, per-class accuracy, and predictions on the testing dataset to evaluate the reproducibility. For the overall accuracy, we compare the differences between each pair of original model and validation model and report the maximum differences and the standard deviation. For per-class accuracy, we first compare the differences of each label between the original model and validation model and extract the maximum difference across all labels and the standard deviation. For the prediction results, we first extract the inconsistent prediction results between the original model and the validation model, then we list the maximum of inconsistent predictions and the standard deviation.

In addition, for each experiment, we also collect the running time for each of the above experiment to assess the runtime overhead incurred by our approach. We only focus on the experiments conducted on GPU, as GPU-based experiments are executed on a physical machine. CPU-based experiments are conducted on a virtual machine in the cloud environment, which can introduce large variances caused by the underlying cloud platform~\cite{ScheunerWWW15}. 
%For the experiments on GPU, we compare the running time between each setup. 
For example, comparing the time of G1 and G3 could reveal the performance impact on enabling deterministic patch for GPU. Comparing the time of G3 and G5 could reveal the overhead introduced through record-and-replay technique. To statistically compare the time differences, we perform the non-parametric Wilcoxon rank-sum test (WSR). To assess the magnitude of the time differences among different setups, we also calculate the effect size using Cliff's Delta~\cite{romano2006appropriate}. 

Finally, as our approach also stores additional data (e.g., the recorded random profile during the store-and-replay phase), we evaluate the storage overhead brought by our approach by comparing the size of DL models with the size of random profiles.

\subsection{Evaluation Analysis and Results}
\label{sec:evaluate}

Here we evaluate if the studied models are reproducible after applying our approach. Then we study the time and storage overhead associated with our approach.

\noindent\textbf{Reproducibility by applying our approach}. The results show that, the six open source models can be successfully reproduced by applying our approach with default settings. In other words, all the predictions are consistent between the target model and the reproduced model. The default record-and-replay technique intercepts two types of randomness introducing system calls (i.e., the read of \texttt{/dev/urandom} and \texttt{getrandom}). The default patch is the version 0.3.0 of tensorflow-determinism released in PyPI for TensorFlow 1.14. For TensorFlow 2.1, we need to set the environment variable \texttt{TF\_CUDNN\_DETERMINISTIC} to "true".
The results demonstrate the effectiveness of our approach on training reproducible DL models. % In addition, in the Setup C2, C4, G4, and G8, with the same seeds and GPU patch, the DL models are also reproducible.

Unfortunately, ModelX under such default setup cannot be reproduced. 
%Our experience on reproducing ModelX further reveals the benefits of our profile-and-patch technique.
While applying our approach, during the profiling and diagnosing phase, we found one library function (\texttt{unsorted\_segment\_s\\um}) invoked from ModelX, which cannot be mitigated by the default patch.
We carefully examined the solutions described in~\cite{nvidiadeterminism} and discovered an experimental patch that could resolve this issue.
We applied the experimental patch along with the record-and-replay technique and are able to achieve reproducibility for ModelX, i.e., all the predictions are consistent.

\noindent{\textbf{Overhead}}. We evaluate the overall time overhead incurred by our approach by comparing training time between the setup without seed, record-and-replay, and patch against the setup with record-and-replay and patch (a.k.a., our approach).  We only compare the training time among open source models, as ModelX adopts the early-stopping mechanism as described above (Section~\ref{sec:casestudy}). As shown in Table~\ref{tab:time_overhead}, training takes longer when applying our approach than the setups without. This is mainly because patched functions adopt deterministic operations, which do not leverage operations (e.g., \texttt{atomicAdd}) that support parallel computation. The time overhead ranges from 24\% to 114\% in our experiments. Although our approach makes training on GPU slower, compared with training on CPUs, training on GPU with our approach is still much faster (e.g., training WRN-28-10 on CPU takes more than 7 days).
We further evaluate the time overhead brought by patching and record-and-replay alone. We compare the setup with patching enabled against the setups without it (e.g., G1 vs. G3). We also compare the setup with record-and-replay, patching enabled with the setup with patching only (e.g., G3 vs. G5). The results show that the record-and-replay technique does not introduce statistical significant overhead ($p-value > 0.05$). In other words, patching is the main reason that our approach introduces the time overhead.

\begin{table}[h]
	\centering
	\caption{Comparing the time and storage overhead. Time(O) represents the average training time (in hours) for original setup, and Time(R) represents the average training time (in hours) for the setup using our approach (Time(R)). The time is italicized if p-value is \textless 0.001 and the effect size is large with (*). RP represents for Random Profile.}
%\begin{tabular}{lllll}
%	\toprule
%	Model & Time(O) & Time(R) & p-values (WRS) & Cliff's Delta(d) \\
%	\midrule
%	LeNet-1 &  0.017       &   0.023      &   \textless 0.001  & -1 (large) \\
%	LeNet-4 &  0.019       &   0.027      &    \textless 0.001  & -1 (large) \\
%	LeNet-5 & 0.021 & 0.028 &   \textless 0.001  & -1 (large) \\
%	ResNet-38 & 1.243 & 1.561 &   \textless 0.001  & -1 (large) \\
%	ResNet-56 & 1.752 & 2.179&   \textless 0.001  & -1 (large) \\
%	WRN-28-10 &7.08 & 14.979&   \textless 0.001  & -1 (large) \\
%	\bottomrule
%\end{tabular}
\begin{tabular}{llll}
	\toprule
	\textbf{Model} & \textbf{Time(O)/Time(R)} & \textbf{Model Size} & \textbf{RP Size} \\
	\midrule
	LeNet-1 &  \textit{0.017/0.023} (*)   & 35 KB      & 13 KB    \\
	LeNet-4 &  \textit{0.019/0.027} (*)     & 224 KB     & 13 KB    \\
	LeNet-5 & \textit{0.021/0.028} (*)   & 267 KB     & 13 KB  \\
	ResNet-38 & \textit{1.243/1.561} (*) & 4.8 MB     & 13 KB   \\
	ResNet-56 & \textit{1.752/2.179} (*)&7.6 MB     & 13 KB  \\
	WRN-28-10 &\textit{7.08/14.979} (*)  & 279 MB     & 13 KB \\
	ModelX & - & 675 KB     & 38 KB \\
	\bottomrule
\end{tabular}
\label{tab:time_overhead}
\end{table}

Table~\ref{tab:time_overhead} also shows the average size of trained DL models and the random profiles.
The absolute storage sizes of the random profile are very small, ranging between 13 KB to 38 KB depending on the DL models. %The biggest random profile is generated while training ModelX, which is only 38KB. The size of other random profiles is just 13 KB. 
Compared to the size of the model, the biggest model is WRN-28-10 (279 MB). The random profile is only 0.005\% of the model in terms of the size. When the model is less complex (e.g., LeNet-1), the additional cost becomes more prominent. In LeNet-1, the random profile incurs 37\% additional storage. However, the total storage size when combining the model and the random profile for LetNet-1 is less than 50 KB, which is acceptable under most of the use cases.
%\begin{table}[h]
%	\centering
%	\caption{The storage costs of record-and-replay\jack{merge this with table 3}}
%	\begin{tabular}{lcc}
%		\toprule
%		& Model Size & Random Profile Size \\
%		\midrule
%		LeNet1    & 35 KB      & 13 KB                \\
%		LeNet4    & 224 KB     & 13 KB                \\
%		LeNet5    & 267 KB     & 13 KB                \\
%		ResNet38  & 4.8 MB     & 13 KB                \\
%		ResNet56  & 7.6 MB     & 13 KB                \\
%		WRN-28-10 & 279 MB     & 13 KB                 \\
%		ModelX & 675 KB     & 38 KB              \\
%		\bottomrule
%	\end{tabular}
%	\label{tab:space_overhead}
%\end{table}

\hypobox{\textbf{Summary:} Case study results show that our approach can successfully reproduce all the studied DL models. Patching (i.e., replace non-deterministic operations from hardware with deterministic ones) incurs large time overhead as the trade-off for ensuring deterministic behavior. The record-and-replay technique does not incur additional time overhead in the training process with very small additional storage sizes.}

\section{Discussions}
\label{sec:discussion}
In this section, we conduct the variance analysis and discuss the lessons learnt when applying our approach.

\subsection{Variance Analysis}

To measure the variances introduced by different sources of non-determinism, we compare the evaluation metrics among different setups. Such analysis demonstrates the variances between our approach with the state-of-the-art techniques towards reproducing DL models.
For example, variances caused by software are analyzed by comparing the evaluation metrics between each pair in G3, where patching is enabled to eliminate hardware non-determinism (i.e., the approach proposed by~\cite{nvidiadeterminism}). 
To measure the variances caused by hardware, we compare the evaluation metrics between each pair in G2 or G7, where the random seeds are preset to eliminate software randomness (i.e., the approach proposed by~\cite{PhamASE20}).
In addition to measuring the software variance and hardware variance, which result from applying two state-of-the-art techniques, we also show the variances incurred from the original setup with no preset seed, record-and-replay not enabled, and patching not enabled. The results of our approach, which incurs zero variances, are also listed in the table.
%To measure the variance caused by different versions of TensorFlow, we compare the differences between G4 and G9.
%The detailed results are shown in the paper due to page limitation and we include the collected evaluation metrics of every run in our replication package.
% Here we briefly discuss the highlights of our analysis.

% Table~\ref{tab:variance_overall_acc} shows our results. All three sources of non-determinism have impact on the model performance.
The detailed results are shown in Table~\ref{tab:variance_overall_acc}. We only include the results for the six open source projects due to confidentiality reasons. Three evaluation metrics are used: overall accuracy, per-class accuracy, and the consistency of predictions. 
For each type of metric, we calculate the maximum differences and the standard deviations of the differences.

% For all the DL models, the largest variance is from the original setup. 
For example, for ResNet-38, the largest variance of overall accuracy in the original setup is 2.0\%, while the largest variances introduced by software randomness and hardware non-determinism are 1.4\% and 1.2\%, respectively. 
For per-class accuracy, the largest variance in the original setup is 10.1\%, while the largest variances introduced by software randomness and hardware non-determinism are 6.8\% and 4.9\%.
For predictions, the largest number of inconsistent predictions in the original setup is 219, while the largest number of inconsistent predictions caused by software randomness and hardware non-determinism are 216 and 209, respectively.

In summary, the variances caused by software are generally larger than those caused by hardware, yet the variances caused by hardware are not negligible and need to be controlled in order to train reproducible DL models. The results demonstrate the importance and effectiveness of applying our approach for training reproducible DL models, as our approach is the only one that does not introduce any variances.

% For example, for WRN-28-10, the asset variance incur 1.1\% in overall accuracy, 26.1\% in per-class accuracy, and cause as high as 460 differences in the prediction results. Without controlling any sources of non-determinism, the variance is the largest as shown the columns of overall variance. For each type of variance, we find that software variance usually is the largest except for WRN-28-10 on the overall accuracy.

%\begin{comment}
\begin{table*}[]
	\centering
	\caption{Comparing variances between our approach and the state-of-the-art techniques. Software variance refers to the technique for only controlling hardware non-determinism~\cite{nvidiadeterminism}. Hardware variance refers to the technique for only controlling software randomness~\cite{PhamASE20}. Original variance refers to the variance caused by the original setup.}
	\begin{tabular}{c|lllllllll}
		\toprule
		\multicolumn{1}{l}{}            &           & \multicolumn{2}{l}{\textbf{Our Variance} } & \multicolumn{2}{l}{\textbf{Software Variance} } & \multicolumn{2}{l}{\textbf{Hardware Variance}} & \multicolumn{2}{l}{\textbf{Original Variance}} \\
		\multicolumn{1}{l}{}            &           & Diff               & SDev               & Diff                 & SDev                & Diff                 & SDev                & Diff                & SDev               \\
		\midrule
		\multirow{6}{*}{Overall acc.}   & LeNet1    &   0               &          0          &   0.8\%                   &   0.2\%                  &       0               &   0                  &   1.7\%                  &  0.3\%                  \\
		& LeNet4    &  0                  & 0                  &0.7\%                      & 0.1\%                     & 0                     &0                     & 0.8\%                    &   0.1\%                 \\
		& LeNet5    &  0                  & 0                   & 0.5\%                     & 0.1\%                    & 0                     &  0                   &  0.5\%                   &0.1\%                    \\
		& ResNet38  & 0                & 0                & 1.4\%                  & 0.3\%                 & 1.2\%                  & 0.3\%                 & 2.0\%                 & 0.4\%                \\
		& ResNet56  & 0                & 0                & 1.2\%                  & 0.3\%                 & 0.8\%                  & 0.2\%                & 1.7\%                & 0.3\%                \\
		& WRN-28-10 & 0                & 0               & 1.4\%                  & 0.4\%                 & 1.7\%                 & 0.5\%                 & 2.4\%                 & 0.5\%                \\
		\hline
		\multirow{6}{*}{Per-class acc.} & LeNet1    &     0               &  0                  &   3.7\%                  &0.8\%                     &    0                  &    0                 &  4.8\%                  &       1.2\%             \\
		& LeNet4    &     0             & 0                    & 1.7\%                     & 0.3\%                    & 0                     & 0                     & 3.0\%                    &  0.6\%                  \\
		& LeNet5    &    0                & 0                   &      2.3\%                & 0.4\%                    &  0                    &     0                &  2.5\%                  & 0.5\%                   \\
		& ResNet38  & 0                & 0                & 6.8\%                  & 1.2\%                 & 4.9\%                 & 0.9\%                 & 10.1\%                & 1.9\%                \\
		& ResNet56  & 0                & 0                & 6.8\%                  & 1.1\%                 & 5.3\%                  & 0.8\%                 & 10.5\%                & 1.9\%                \\
		& WRN-28-10 & 0              & 0                & 35.0\%               & 5.0\%                 & 25.0\%                 & 3.0\%                 & 40.9\%                & 7.8\%                \\
		\hline
		\multirow{6}{*}{Predictions}    & LeNet1    &      0              &    0                &  48                   &  14.1                   &     0                 &   0                  & 50                    &  17.04                  \\
		& LeNet4    &    0                &  0                 &  31                    &   3.8                  &    0                  &     0                &     29                &  3.8                  \\
		& LeNet5    &    0                & 0                    & 28                    &   3.5                  & 0                      & 0                    &  26                   &     3.5               \\
		& ResNet38  & 0                & 0               & 216                  & 10.1                & 209                  & 11.3                & 219                 & 10.7                  \\
		& ResNet56  & 0                & 0               & 198                 & 8.6                & 188                  & 8.8                & 198                 & 8.0                  \\
		& WRN-28-10 & 0               & 0               & 485                 & 18.0                & 453                  & 12.3                & 542                & 18.7                 \\
%		\hline
%		MAE loss & Model X & - & - & 0.008 & 0.003 & \textbf{0.029} & 0.009 & 0.035 & 0.006 \\
%		\hline
%		Convergence epoch & Model X & - & - & 17 & 5.3 & 12 & 4.6 & 32 & 7.7 \\
		\bottomrule
	\end{tabular}
	
	\label{tab:variance_overall_acc}
\end{table*}

\subsection{Generalizability in other DL frameworks}

Other than the DL framework studied in Section~\ref{sec:evaluate}, we have also applied our approach on another popular DL framework, PyTorch. Experiment results show that for common models such as LeNet-5 and ResNet-56 with PyTorch version 1.7, our approach can work out of the box. % Through our analysis, the software-related randomness of these two PyTorch-based DL models are attributed solely to \texttt{/dev/urandom}. For GPU patching,\jack{same or different patch? } according to the Nvidia documentation~\cite{nvidiadeterminism}, PyTorch supports using an API \texttt{use\_deterministic\_algorithms} for replacing non-deterministic functions with deterministic ones. %The results show that our approach can be generalizable with other DL framework with little code alteration. 
In the future, we also plan to experiment our approach on more DL frameworks and more DL models across different tasks. %(e.g., MXNet~\cite{mxnet} and PaddlePaddle~\cite{paddlepaddle}).   

\subsection{Documentations on DL Models}

%\boyuan{Modified shown below}\jack{no, it's not on data sheet. But you should talk more on model cards: (0) overview of model cards, its object and structure/content (1) more rigrious way (e.g., the keras dependecy case) or how data is splitting for train/dev/test, (2) codify and co-evolve such things with code, data and documentation as things may constantly changing}
Mitchell et al.~\cite{MitchellFAT19} proposed \textsf{Model Cards} to document ML models.
A typical model card includes nine sections (e.g., Model Details and Intended Use), each of which contains a list of relevant information. For example, in the Model Details section, it suggests that the ``Information about training algorithms, parameters, fairness constraints or other applied approaches, and features'' should be accompanied with released models. 
Such a practice would help other researchers or practitioners to evaluate if the models can be reproduced. However, the current practice would still miss certain details. We share our experience below to demonstrate this point.

TensorFlow and Keras are two of the most widely used DL frameworks.
Keras is a set of high level APIs designed for simplicity and usability for both software engineers and DL researchers, while TensorFlow offers more low level operations and is more flexible to design and implement complex network structures.
There are two ways of using Keras and TensorFlow in DL training.
The first way is to import Keras and TensorFlow separately by first calling \texttt{import keras} and then verify if the backend of Keras is TensorFlow.
If yes, TensorFlow can be imported by \texttt{import tensorflow}. 
This way is referred to as \textit{Keras\_first}.
The second way is to directly use the Keras API within TensorFlow by first importing TensorFlow.
Then we use another import statement \texttt{from tensorflow import keras}. This way is referred to as \textit{TF\_first}.
We conduct experiments to evaluate if the two different usage of APIs have an impact on training reproducible DL models.
As a result, the following findings are presented:

\begin{itemize}[leftmargin=*]
	\item When training on CPUs, using \textit{Keras\_first} will lead to unreproducible results even after mitigating all the sources of non-determinism. This issue can be reproduced by using various Keras version from 2.2.2 to 2.2.5.
	On the contrary, using \textit{TF\_first} with the same setting will yield reproducible results. This issue does not exist in training on GPUs.
	
	\item While training with Keras version 2.3.0 and above, we are able to reproduce the results both for \textit{Keras\_first} and \textit{TF\_first} using our approach.
	However, the DL models trained using \textit{Keras\_first} and \textit{TF\_first} are not consistent with each other.
\end{itemize}

Both findings have been submitted as issue reports to the official Keras development team who suggested us to use newer versions of Keras instead~\cite{kerasissue1,kerasissue2}.
The findings highlight that not only the versions of dependencies, but also how the dependent software packages are used can impact the reproducibility of DL models. Unfortunately, existing DL model documentation frameworks like \textsf{Model cards}~\cite{MitchellFAT19} do not specify how the software dependencies should be described. Hence, we suggest ML practitioners look into the approach adopted for traditional software projects like software bills of materials (SBOM)~\cite{SBOM:2020}
% or software package data exchange (SPDX)~\cite{spdx} 
for rigorously specifying software dependencies.

\section{Guideline}
\label{sec:guideline}

%\jack{This needs to be a section of its own right after evaluation section. mention as 5 steps: (1) model cards for documention, (2) assert management tool aspect like version control tools, and experimental managment tools, (3) evaluation criteria under different context, (4) software store/replay and hardware patch to control non-determinism, (5) if still not reproduciable, check if random function changes or some types of API not supported}\boyuan{changed}

In this section, we propose a guideline for researchers and practitioners who are interested in constructing reproducible DL models. Our guideline consists of five steps:

\begin{enumerate}[leftmargin=*]
	\item Use documentation frameworks such as \texttt{Model Cards} to document the details such as model training. Consider leveraging SBOM to document software dependencies. Ensure the documentation co-evolves with the model development process.
	
	\item Use asset management tools such as DVC~\cite{dvc} and MLflow~\cite{mlflow} to manage the experimental assets used during training process. To mitigate the risks of introducing non-determinism from assets, we suggest using virtualization techniques to provide a complete runtime environment.%  Version management tools (e.g., Git) should be considered to manage the evolution of assets. 
	
	\item Use and document the appropriate evaluation criteria depending on the domain of the DL models. Some of these metrics (e.g., evaluation metrics ) may be domain specific, whereas other metrics (e.g., process metrics) are general. %For example, classification and regression tasks should adopt different performance metrics for verifying reproducibility.
	
	\item Randomness in the software and non-determinism from hardware are two of the main challenges preventing the reproducibility of DL models. Use record-and-replay technique to mitigate sources of randomness in the software when presetting seed is not preferred. Use patching to mitigate the non-determinism from hardware if the overhead is acceptable. %\boyuan{done}\jack{mention on the store/replay framework and patches for setup}
	% Setting seeds beforehand or using our record and replay mechanism could control the randomness completely. While setting seed is not preferable in certain contexts, we suggest using our record and replay mechanism as it incurs minimal intrusion to the process. The random profile should be kept associated with each training for further validation and auditing process.
	
	\item If DL models are still not reproducible by applying our approach, double check if the list of system calls which introduce randomness changes or if the deterministic operations are not currently supported by the hardware libraries. Document the unsupported non-deterministic operations and search for alternative operations on the same operation. %Continuously track the status of such operations from the official Nvidia repository.
	
	% If yes, repeat the previous steps and try again. If no, document the list of unsupported operations, so that you can file the continuously track their status from the official Nvdia website.\jack{double check this}
	
\end{enumerate}
\section{Threats to Validity}
\label{sec:threats}
%In this section, we discuss the threats to validity.
%\subsubsection*{External Validity}

\noindent\textbf{External Validity.} Currently, we focus on DL training using Python along with TensorFlow and Keras framework under Linux. We are currently working on extending our approach to support DL models developed in other DL frameworks and additional operating systems. In addition, we have applied our approaches on two popular domains of DL: classification and regression tasks. We plan to investigate other tasks such as Natural Language Processing and Reinforcement Learning. %Furthermore, current approach only supports Linux, we plan to extend our approach in other operating systems (e.g., Windows).
GPUs and CPUs are common and widely adopted hardware for DL training. Hence, in this paper, we choose to focus on evaluating the DL training on GPUs and CPUs. However, DL training on other hardware such as TPU and edge devices also might encounter reproducibility issues. We believe the idea of our approach can be applied in these contexts as well. Future work is welcomed to extend our approach to different platforms. 

%\subsubsection*{Internal Validity}
%\jack{these things do not seem to be internal valdity issues to me. i feel it's generally about confounding factors, here we have three things: (1) asset, (2)sw, and (3) hw, we have come up approaches for these different aspects, and in the case study inspected their impact separately by keeping the others the same}\boyuan{added}

\noindent\textbf{Internal Validity.} When measuring the variances incurred by different sources of non-determinism, we control the other confounding factors to ensure internal validity. For example, when measuring the overall accuracy variance caused by randomness in software, we only compare the runs with patching enabled and with the same dependencies. In addition, in our evaluation, we repeat the model training process for at least 16 times for each setup to observe the impact of different non-deterministic factors. % Hence, the chances of implementation bugs in these DL models are very rare..

% When the generated model have the exact same accuracy on the testing dataset, it did not mean that the model binary files are exactly the same. Similar to the traditional software system, the binary equivalences of model files is an important and interesting topic and worth future investigation.

%\subsubsection*{Construct Validity}
\noindent\textbf{Construct Validity.} The implementation code for the DL models used in our case studies has been careful reviewed by previous researchers~\cite{PhamASE20,GuoASE19,GuIFIP21,MaFSE18}. Our record-and-replay technique for controlling the software factors work when low level random functions are dynamically linked and invoked. % \jack{I am not sure if the remaining part is needed?}However, the API hook mechanism does not work when the random functions are within a static library and not dynamically linked during execution. Additional efforts need to be made to convert the static libraries to dynamic libraries. How to control random functions in static libraries remains an open problem. 
\section{Conclusions}
\label{sec:conclusion}
Reproducibility is a rising concern in AI, especially in DL. Prior practices and research mainly focus on mitigating the sources of non-determinism separately without a systematic approach and thorough evaluation. In this paper, we propose a systematic approach to reproducing DL models through controlling the software and hardware non-determinism. 
%We make sure the DL experiments have consistent dependencies for tackling the environment factors. We propose a novel record and replay technique to record the randomness states in the DL experiments for software factors. We use profiling based approach to systematically examine the hardware related non-determinism. 
Case studies on six open source and one commercial DL models show that all the models can be successfully reproduced by leveraging our approach. 
In addition, we present a guideline for training reproducible DL models and describe some of the lessons learned based on our experience of applying our approach in practice. Last, we provide a replication package~\cite{replicationpackage} to facilitate reproducibility of our study.

\bibliographystyle{ACM-Reference-Format}
\bibliography{icse2022}


\begin{thebibliography}{69}

%%% the Web address.

\ifx \showCODEN    \undefined \def \showCODEN     #1{\unskip}     \fi
\ifx \showDOI      \undefined \def \showDOI       #1{#1}\fi
\ifx \showISBNx    \undefined \def \showISBNx     #1{\unskip}     \fi
\ifx \showISBNxiii \undefined \def \showISBNxiii  #1{\unskip}     \fi
\ifx \showISSN     \undefined \def \showISSN      #1{\unskip}     \fi
\ifx \showLCCN     \undefined \def \showLCCN      #1{\unskip}     \fi
\ifx \shownote     \undefined \def \shownote      #1{#1}          \fi
\ifx \showarticletitle \undefined \def \showarticletitle #1{#1}   \fi
\ifx \showURL      \undefined \def \showURL       {\relax}        \fi
% invisible to TeX
\providecommand\bibfield[2]{#2}
\providecommand\bibinfo[2]{#2}
\providecommand\natexlab[1]{#1}
\providecommand\showeprint[2][]{arXiv:#2}

\bibitem[\protect\citeauthoryear{??}{mlf}{2021}]%
        {mlflow}
 \bibinfo{year}{2021 (accessed August, 2021)}\natexlab{}.
\newblock \bibinfo{booktitle}{\emph{{An open source platform for the machine
  learning lifecycle}}}.
\newblock
\urldef\tempurl%
\url{https://mlflow.org/}
\showURL{%
\tempurl}


\bibitem[\protect\citeauthoryear{??}{ass}{2021}]%
        {assessmentEurope}
 \bibinfo{year}{2021 (accessed August, 2021)}\natexlab{}.
\newblock \bibinfo{booktitle}{\emph{{Assessment List for Trustworthy Artificial
  Intelligence (ALTAI) for self-assessment}}}.
\newblock
\urldef\tempurl%
\url{https://digital-strategy.ec.europa.eu/en/library/assessment-list-trustworthy-artificial-intelligence-altai-self-assessment}
\showURL{%
\tempurl}


\bibitem[\protect\citeauthoryear{??}{cif}{2021}]%
        {cifar}
 \bibinfo{year}{2021 (accessed August, 2021)}\natexlab{}.
\newblock \bibinfo{booktitle}{\emph{The CIFAR-10 and CIFAR-100 datasets}}.
\newblock
\urldef\tempurl%
\url{https://www.cs.toronto.edu/~kriz/cifar.html}
\showURL{%
\tempurl}


\bibitem[\protect\citeauthoryear{??}{cud}{2021}]%
        {cuda}
 \bibinfo{year}{2021 (accessed August, 2021)}\natexlab{}.
\newblock \bibinfo{booktitle}{\emph{{CUDA Toolkit}}}.
\newblock
\urldef\tempurl%
\url{https://developer.nvidia.com/cuda-toolkit}
\showURL{%
\tempurl}


\bibitem[\protect\citeauthoryear{??}{det}{2021}]%
        {determined-repro}
 \bibinfo{year}{2021 (accessed August, 2021)}\natexlab{}.
\newblock \bibinfo{booktitle}{\emph{{Determined AI Reproducibility}}}.
\newblock
\urldef\tempurl%
\url{https://docs.determined.ai/latest/topic-guides/training/reproducibility.html}
\showURL{%
\tempurl}


\bibitem[\protect\citeauthoryear{??}{dun}{2021}]%
        {duncanGTC19}
 \bibinfo{year}{2021 (accessed August, 2021)}\natexlab{}.
\newblock \bibinfo{booktitle}{\emph{Determinism in Deep Learning (S9911)}}.
\newblock
\urldef\tempurl%
\url{https://developer.download.nvidia.com/video/gputechconf/gtc/2019/presentation/s9911-determinism-in-deep-learning.pdf}
\showURL{%
\tempurl}


\bibitem[\protect\citeauthoryear{??}{ker}{2021a}]%
        {kerasissue1}
 \bibinfo{year}{2021 (accessed August, 2021)}\natexlab{a}.
\newblock \bibinfo{booktitle}{\emph{Inconsistent results when using two styles
  of import statements - Issue 14672}}.
\newblock
\urldef\tempurl%
\url{https://github.com/keras-team/keras/issues/14672}
\showURL{%
\tempurl}


\bibitem[\protect\citeauthoryear{??}{tfm}{2021}]%
        {tfmigrate}
 \bibinfo{year}{2021 (accessed August, 2021)}\natexlab{}.
\newblock \bibinfo{booktitle}{\emph{{Migrate your TensorFlow 1 code to
  TensorFlow 2}}}.
\newblock
\urldef\tempurl%
\url{https://www.tensorflow.org/guide/migrate}
\showURL{%
\tempurl}


\bibitem[\protect\citeauthoryear{??}{mni}{2021}]%
        {mnist}
 \bibinfo{year}{2021 (accessed August, 2021)}\natexlab{}.
\newblock \bibinfo{booktitle}{\emph{The Mnist Database of handwritten digits}}.
\newblock
\urldef\tempurl%
\url{http://yann.lecun.com/exdb/mnist/}
\showURL{%
\tempurl}


\bibitem[\protect\citeauthoryear{??}{com}{2021}]%
        {commonseeds}
 \bibinfo{year}{2021 (accessed August, 2021)}\natexlab{}.
\newblock \bibinfo{booktitle}{\emph{Most common random seeds}}.
\newblock
\urldef\tempurl%
\url{https://www.kaggle.com/residentmario/kernel16e284dcb7}
\showURL{%
\tempurl}


\bibitem[\protect\citeauthoryear{??}{mck}{2021}]%
        {mckinseyreport}
 \bibinfo{year}{2021 (accessed August, 2021)}\natexlab{}.
\newblock \bibinfo{booktitle}{\emph{Notes from the AI Frontier Insights from
  Hundreds of Use Cases}}.
\newblock
\urldef\tempurl%
\url{https://www.mckinsey.com/featured-insights/artificial-intelligence/
  notes-from-the-ai-frontier-applications-and-value-of-deep-learning}
\showURL{%
\tempurl}


\bibitem[\protect\citeauthoryear{??}{dvc}{2021}]%
        {dvc}
 \bibinfo{year}{2021 (accessed August, 2021)}\natexlab{}.
\newblock \bibinfo{booktitle}{\emph{{Open-source Version Control System for
  Machine Learning Projects}}}.
\newblock
\urldef\tempurl%
\url{https://dvc.org/}
\showURL{%
\tempurl}


\bibitem[\protect\citeauthoryear{??}{Eur}{2021}]%
        {EuroRegulation}
 \bibinfo{year}{2021 (accessed August, 2021)}\natexlab{}.
\newblock \bibinfo{booktitle}{\emph{{Proposal for a REGULATION OF THE EUROPEAN
  PARLIAMENT AND OF THE COUNCIL LAYING DOWN HARMONISED RULES ON ARTIFICIAL
  INTELLIGENCE (ARTIFICIAL INTELLIGENCE ACT) AND AMENDING CERTAIN UNION
  LEGISLATIVE ACTS}}}.
\newblock
\urldef\tempurl%
\url{https://eur-lex.europa.eu/legal-content/EN/TXT/?uri=CELEX%3A52021PC0206}
\showURL{%
\tempurl}


\bibitem[\protect\citeauthoryear{??}{pyt}{2021}]%
        {pytorchrandom}
 \bibinfo{year}{2021 (accessed August, 2021)}\natexlab{}.
\newblock \bibinfo{booktitle}{\emph{{Reproducibility in Pytorch}}}.
\newblock
\urldef\tempurl%
\url{https://pytorch.org/docs/stable/notes/randomness.html}
\showURL{%
\tempurl}


\bibitem[\protect\citeauthoryear{??}{nvi}{2021}]%
        {nvidiadeterminism}
 \bibinfo{year}{2021 (accessed August, 2021)}\natexlab{}.
\newblock \bibinfo{booktitle}{\emph{{Tensorflow Determinism}}}.
\newblock
\urldef\tempurl%
\url{https://github.com/NVIDIA/framework-determinism}
\showURL{%
\tempurl}


\bibitem[\protect\citeauthoryear{??}{tfv}{2021}]%
        {tfversion}
 \bibinfo{year}{2021 (accessed August, 2021)}\natexlab{}.
\newblock \bibinfo{booktitle}{\emph{TensorFlow GPU Support}}.
\newblock
\urldef\tempurl%
\url{https://www.tensorflow.org/install/source#gpu}
\showURL{%
\tempurl}


\bibitem[\protect\citeauthoryear{??}{tfr}{2021}]%
        {tfreprorfc}
 \bibinfo{year}{2021 (accessed August, 2021)}\natexlab{}.
\newblock \bibinfo{booktitle}{\emph{{Tensorflow RFC for determinism}}}.
\newblock
\urldef\tempurl%
\url{https://github.com/tensorflow/community/blob/master/rfcs/20210119-determinism.md}
\showURL{%
\tempurl}


\bibitem[\protect\citeauthoryear{??}{tes}{2021}]%
        {testMLGoogle}
 \bibinfo{year}{2021 (accessed August, 2021)}\natexlab{}.
\newblock \bibinfo{booktitle}{\emph{{Testing for Deploying Machine Learning
  Models}}}.
\newblock
\urldef\tempurl%
\url{https://developers.google.com/machine-learning/testing-debugging/pipeline/deploying}
\showURL{%
\tempurl}


\bibitem[\protect\citeauthoryear{??}{mlr}{2021}]%
        {mlreprochecklist}
 \bibinfo{year}{2021 (accessed August, 2021)}\natexlab{}.
\newblock \bibinfo{booktitle}{\emph{{The Machine Learning Reproducibility
  Checklist}}}.
\newblock
\urldef\tempurl%
\url{https://www.cs.mcgill.ca/~jpineau/ReproducibilityChecklist.pdf}
\showURL{%
\tempurl}


\bibitem[\protect\citeauthoryear{??}{ker}{2021b}]%
        {kerasissue2}
 \bibinfo{year}{2021 (accessed August, 2021)}\natexlab{b}.
\newblock \bibinfo{booktitle}{\emph{Unreproducible results when directly import
  keras in CPU environment - Issue 14671}}.
\newblock
\urldef\tempurl%
\url{https://github.com/keras-team/keras/issues/14671}
\showURL{%
\tempurl}


\bibitem[\protect\citeauthoryear{??}{rep}{2022}]%
        {replicationpackage}
 \bibinfo{year}{2022 (accessed Feb, 2022)}\natexlab{}.
\newblock \bibinfo{booktitle}{\emph{The replication package}}.
\newblock
\urldef\tempurl%
\url{https://github.com/nemo9cby/ICSE2022Rep}
\showURL{%
\tempurl}


\bibitem[\protect\citeauthoryear{Abadi, Barham, Chen, Chen, Davis, Dean, Devin,
  Ghemawat, Irving, Isard, Kudlur, Levenberg, Monga, Moore, Murray, Steiner,
  Tucker, Vasudevan, Warden, Wicke, Yu, and Zheng}{Abadi et~al\mbox{.}}{2016}]%
        {AbadiOSDI16}
\bibfield{author}{\bibinfo{person}{Mart{\'{\i}}n Abadi}, \bibinfo{person}{Paul
  Barham}, \bibinfo{person}{Jianmin Chen}, \bibinfo{person}{Zhifeng Chen},
  \bibinfo{person}{Andy Davis}, \bibinfo{person}{Jeffrey Dean},
  \bibinfo{person}{Matthieu Devin}, \bibinfo{person}{Sanjay Ghemawat},
  \bibinfo{person}{Geoffrey Irving}, \bibinfo{person}{Michael Isard},
  \bibinfo{person}{Manjunath Kudlur}, \bibinfo{person}{Josh Levenberg},
  \bibinfo{person}{Rajat Monga}, \bibinfo{person}{Sherry Moore},
  \bibinfo{person}{Derek~Gordon Murray}, \bibinfo{person}{Benoit Steiner},
  \bibinfo{person}{Paul~A. Tucker}, \bibinfo{person}{Vijay Vasudevan},
  \bibinfo{person}{Pete Warden}, \bibinfo{person}{Martin Wicke},
  \bibinfo{person}{Yuan Yu}, {and} \bibinfo{person}{Xiaoqiang Zheng}.}
  \bibinfo{year}{2016}\natexlab{}.
\newblock \showarticletitle{TensorFlow: {A} System for Large-Scale Machine
  Learning}. In \bibinfo{booktitle}{\emph{12th {USENIX} Symposium on Operating
  Systems Design and Implementation, {OSDI} 2016, Savannah, GA, USA, November
  2-4, 2016}}. \bibinfo{publisher}{{USENIX} Association},
  \bibinfo{pages}{265--283}.
\newblock


\bibitem[\protect\citeauthoryear{Amershi, Begel, Bird, DeLine, Gall, Kamar,
  Nagappan, Nushi, and Zimmermann}{Amershi et~al\mbox{.}}{2019}]%
        {AmershiSEIP19}
\bibfield{author}{\bibinfo{person}{Saleema Amershi}, \bibinfo{person}{Andrew
  Begel}, \bibinfo{person}{Christian Bird}, \bibinfo{person}{Robert DeLine},
  \bibinfo{person}{Harald~C. Gall}, \bibinfo{person}{Ece Kamar},
  \bibinfo{person}{Nachiappan Nagappan}, \bibinfo{person}{Besmira Nushi}, {and}
  \bibinfo{person}{Thomas Zimmermann}.} \bibinfo{year}{2019}\natexlab{}.
\newblock \showarticletitle{Software engineering for machine learning: a case
  study}. In \bibinfo{booktitle}{\emph{Proceedings of the 41st International
  Conference on Software Engineering: Software Engineering in Practice, {ICSE}
  {(SEIP)} 2019, Montreal, QC, Canada, May 25-31, 2019}},
  \bibfield{editor}{\bibinfo{person}{Helen Sharp} {and} \bibinfo{person}{Mike
  Whalen}} (Eds.). \bibinfo{publisher}{{IEEE} / {ACM}},
  \bibinfo{pages}{291--300}.
\newblock


\bibitem[\protect\citeauthoryear{Amodei, Ananthanarayanan, Anubhai, Bai,
  Battenberg, Case, Casper, Catanzaro, Chen, Chrzanowski, Coates, Diamos,
  Elsen, Engel, Fan, Fougner, Hannun, Jun, Han, LeGresley, Li, Lin, Narang, Ng,
  Ozair, Prenger, Qian, Raiman, Satheesh, Seetapun, Sengupta, Wang, Wang, Wang,
  Xiao, Xie, Yogatama, Zhan, and Zhu}{Amodei et~al\mbox{.}}{2016}]%
        {AmodeiICML16}
\bibfield{author}{\bibinfo{person}{Dario Amodei}, \bibinfo{person}{Sundaram
  Ananthanarayanan}, \bibinfo{person}{Rishita Anubhai},
  \bibinfo{person}{Jingliang Bai}, \bibinfo{person}{Eric Battenberg},
  \bibinfo{person}{Carl Case}, \bibinfo{person}{Jared Casper},
  \bibinfo{person}{Bryan Catanzaro}, \bibinfo{person}{Jingdong Chen},
  \bibinfo{person}{Mike Chrzanowski}, \bibinfo{person}{Adam Coates},
  \bibinfo{person}{Greg Diamos}, \bibinfo{person}{Erich Elsen},
  \bibinfo{person}{Jesse~H. Engel}, \bibinfo{person}{Linxi Fan},
  \bibinfo{person}{Christopher Fougner}, \bibinfo{person}{Awni~Y. Hannun},
  \bibinfo{person}{Billy Jun}, \bibinfo{person}{Tony Han},
  \bibinfo{person}{Patrick LeGresley}, \bibinfo{person}{Xiangang Li},
  \bibinfo{person}{Libby Lin}, \bibinfo{person}{Sharan Narang},
  \bibinfo{person}{Andrew~Y. Ng}, \bibinfo{person}{Sherjil Ozair},
  \bibinfo{person}{Ryan Prenger}, \bibinfo{person}{Sheng Qian},
  \bibinfo{person}{Jonathan Raiman}, \bibinfo{person}{Sanjeev Satheesh},
  \bibinfo{person}{David Seetapun}, \bibinfo{person}{Shubho Sengupta},
  \bibinfo{person}{Chong Wang}, \bibinfo{person}{Yi Wang},
  \bibinfo{person}{Zhiqian Wang}, \bibinfo{person}{Bo Xiao},
  \bibinfo{person}{Yan Xie}, \bibinfo{person}{Dani Yogatama},
  \bibinfo{person}{Jun Zhan}, {and} \bibinfo{person}{Zhenyao Zhu}.}
  \bibinfo{year}{2016}\natexlab{}.
\newblock \showarticletitle{Deep Speech 2 : End-to-End Speech Recognition in
  English and Mandarin}. In \bibinfo{booktitle}{\emph{Proceedings of the 33nd
  International Conference on Machine Learning, {ICML} 2016, New York City, NY,
  USA, June 19-24, 2016}} \emph{(\bibinfo{series}{{JMLR} Workshop and
  Conference Proceedings})}.
\newblock


\bibitem[\protect\citeauthoryear{Barrak, Eghan, and Adams}{Barrak
  et~al\mbox{.}}{2021}]%
        {BarrakSANER21}
\bibfield{author}{\bibinfo{person}{Amine Barrak}, \bibinfo{person}{Ellis~E.
  Eghan}, {and} \bibinfo{person}{Bram Adams}.} \bibinfo{year}{2021}\natexlab{}.
\newblock \showarticletitle{On the Co-evolution of {ML} Pipelines and Source
  Code - Empirical Study of {DVC} Projects}. In \bibinfo{booktitle}{\emph{28th
  {IEEE} International Conference on Software Analysis, Evolution and
  Reengineering, {SANER} 2021, Honolulu, HI, USA, March 9-12, 2021}}.
  \bibinfo{publisher}{{IEEE}}, \bibinfo{pages}{422--433}.
\newblock


\bibitem[\protect\citeauthoryear{Brundage, Avin, Wang, Belfield, Krueger,
  Hadfield, Khlaaf, Yang, Toner, Fong, Maharaj, Koh, Hooker, Leung, Trask,
  Bluemke, Lebensbold, O'Keefe, Koren, Ryffel, Rubinovitz, Besiroglu, Carugati,
  Clark, Eckersley, de~Haas, Johnson, Laurie, Ingerman, Krawczuk, Askell,
  Cammarota, Lohn, Krueger, Stix, Henderson, Graham, Prunkl, Martin, Seger,
  Zilberman, h{\'{E}}igeartaigh, Kroeger, Sastry, Kagan, Weller, Tse, Barnes,
  Dafoe, Scharre, Herbert{-}Voss, Rasser, Sodhani, Flynn, Gilbert, Dyer, Khan,
  Bengio, and Anderljung}{Brundage et~al\mbox{.}}{2020}]%
        {BrundageArxiv20}
\bibfield{author}{\bibinfo{person}{Miles Brundage}, \bibinfo{person}{Shahar
  Avin}, \bibinfo{person}{Jasmine Wang}, \bibinfo{person}{Haydn Belfield},
  \bibinfo{person}{Gretchen Krueger}, \bibinfo{person}{Gillian~K. Hadfield},
  \bibinfo{person}{Heidy Khlaaf}, \bibinfo{person}{Jingying Yang},
  \bibinfo{person}{Helen Toner}, \bibinfo{person}{Ruth Fong},
  \bibinfo{person}{Tegan Maharaj}, \bibinfo{person}{Pang~Wei Koh},
  \bibinfo{person}{Sara Hooker}, \bibinfo{person}{Jade Leung},
  \bibinfo{person}{Andrew Trask}, \bibinfo{person}{Emma Bluemke},
  \bibinfo{person}{Jonathan Lebensbold}, \bibinfo{person}{Cullen O'Keefe},
  \bibinfo{person}{Mark Koren}, \bibinfo{person}{Theo Ryffel},
  \bibinfo{person}{J.~B. Rubinovitz}, \bibinfo{person}{Tamay Besiroglu},
  \bibinfo{person}{Federica Carugati}, \bibinfo{person}{Jack Clark},
  \bibinfo{person}{Peter Eckersley}, \bibinfo{person}{Sarah de Haas},
  \bibinfo{person}{Maritza Johnson}, \bibinfo{person}{Ben Laurie},
  \bibinfo{person}{Alex Ingerman}, \bibinfo{person}{Igor Krawczuk},
  \bibinfo{person}{Amanda Askell}, \bibinfo{person}{Rosario Cammarota},
  \bibinfo{person}{Andrew Lohn}, \bibinfo{person}{David Krueger},
  \bibinfo{person}{Charlotte Stix}, \bibinfo{person}{Peter Henderson},
  \bibinfo{person}{Logan Graham}, \bibinfo{person}{Carina Prunkl},
  \bibinfo{person}{Bianca Martin}, \bibinfo{person}{Elizabeth Seger},
  \bibinfo{person}{Noa Zilberman}, \bibinfo{person}{Se{\'{a}}n~{\'{O}}
  h{\'{E}}igeartaigh}, \bibinfo{person}{Frens Kroeger}, \bibinfo{person}{Girish
  Sastry}, \bibinfo{person}{Rebecca Kagan}, \bibinfo{person}{Adrian Weller},
  \bibinfo{person}{Brian Tse}, \bibinfo{person}{Elizabeth Barnes},
  \bibinfo{person}{Allan Dafoe}, \bibinfo{person}{Paul Scharre},
  \bibinfo{person}{Ariel Herbert{-}Voss}, \bibinfo{person}{Martijn Rasser},
  \bibinfo{person}{Shagun Sodhani}, \bibinfo{person}{Carrick Flynn},
  \bibinfo{person}{Thomas~Krendl Gilbert}, \bibinfo{person}{Lisa Dyer},
  \bibinfo{person}{Saif Khan}, \bibinfo{person}{Yoshua Bengio}, {and}
  \bibinfo{person}{Markus Anderljung}.} \bibinfo{year}{2020}\natexlab{}.
\newblock \showarticletitle{Toward Trustworthy {AI} Development: Mechanisms for
  Supporting Verifiable Claims}.
\newblock \bibinfo{journal}{\emph{CoRR}}  \bibinfo{volume}{abs/2004.07213}
  (\bibinfo{year}{2020}).
\newblock
\showeprint[arxiv]{2004.07213}
\urldef\tempurl%
\url{https://arxiv.org/abs/2004.07213}
\showURL{%
\tempurl}


\bibitem[\protect\citeauthoryear{Chetlur, Woolley, Vandermersch, Cohen, Tran,
  Catanzaro, and Shelhamer}{Chetlur et~al\mbox{.}}{2014}]%
        {ChetlurArxiv14}
\bibfield{author}{\bibinfo{person}{Sharan Chetlur}, \bibinfo{person}{Cliff
  Woolley}, \bibinfo{person}{Philippe Vandermersch}, \bibinfo{person}{Jonathan
  Cohen}, \bibinfo{person}{John Tran}, \bibinfo{person}{Bryan Catanzaro}, {and}
  \bibinfo{person}{Evan Shelhamer}.} \bibinfo{year}{2014}\natexlab{}.
\newblock \showarticletitle{cuDNN: Efficient Primitives for Deep Learning}.
\newblock \bibinfo{journal}{\emph{CoRR}}  \bibinfo{volume}{abs/1410.0759}
  (\bibinfo{year}{2014}).
\newblock
\showeprint[arxiv]{1410.0759}
\urldef\tempurl%
\url{http://arxiv.org/abs/1410.0759}
\showURL{%
\tempurl}


\bibitem[\protect\citeauthoryear{Colas, Sigaud, and Oudeyer}{Colas
  et~al\mbox{.}}{2018}]%
        {ColasArxiv18}
\bibfield{author}{\bibinfo{person}{C{\'{e}}dric Colas},
  \bibinfo{person}{Olivier Sigaud}, {and} \bibinfo{person}{Pierre{-}Yves
  Oudeyer}.} \bibinfo{year}{2018}\natexlab{}.
\newblock \showarticletitle{How Many Random Seeds? Statistical Power Analysis
  in Deep Reinforcement Learning Experiments}.
\newblock \bibinfo{journal}{\emph{CoRR}}  \bibinfo{volume}{abs/1806.08295}
  (\bibinfo{year}{2018}).
\newblock
\showeprint[arxiv]{1806.08295}
\urldef\tempurl%
\url{http://arxiv.org/abs/1806.08295}
\showURL{%
\tempurl}


\bibitem[\protect\citeauthoryear{Esteva, Robicquet, Ramsundar, Kuleshov,
  DePristo, Chou, Cui, Corrado, Thrun, and Dean}{Esteva et~al\mbox{.}}{2019}]%
        {esteva2019guide}
\bibfield{author}{\bibinfo{person}{Andre Esteva}, \bibinfo{person}{Alexandre
  Robicquet}, \bibinfo{person}{Bharath Ramsundar}, \bibinfo{person}{Volodymyr
  Kuleshov}, \bibinfo{person}{Mark DePristo}, \bibinfo{person}{Katherine Chou},
  \bibinfo{person}{Claire Cui}, \bibinfo{person}{Greg Corrado},
  \bibinfo{person}{Sebastian Thrun}, {and} \bibinfo{person}{Jeff Dean}.}
  \bibinfo{year}{2019}\natexlab{}.
\newblock \showarticletitle{A guide to deep learning in healthcare}.
\newblock \bibinfo{journal}{\emph{Nature medicine}} \bibinfo{volume}{25},
  \bibinfo{number}{1} (\bibinfo{year}{2019}), \bibinfo{pages}{24--29}.
\newblock


\bibitem[\protect\citeauthoryear{Gebru, Morgenstern, Vecchione, Vaughan,
  Wallach, III, and Crawford}{Gebru et~al\mbox{.}}{2018}]%
        {GebruDatasheet18}
\bibfield{author}{\bibinfo{person}{Timnit Gebru}, \bibinfo{person}{Jamie
  Morgenstern}, \bibinfo{person}{Briana Vecchione},
  \bibinfo{person}{Jennifer~Wortman Vaughan}, \bibinfo{person}{Hanna~M.
  Wallach}, \bibinfo{person}{Hal~Daum{\'{e}} III}, {and} \bibinfo{person}{Kate
  Crawford}.} \bibinfo{year}{2018}\natexlab{}.
\newblock \showarticletitle{Datasheets for Datasets}.
\newblock \bibinfo{journal}{\emph{CoRR}}  \bibinfo{volume}{abs/1803.09010}
  (\bibinfo{year}{2018}).
\newblock
\showeprint[arxiv]{1803.09010}
\urldef\tempurl%
\url{http://arxiv.org/abs/1803.09010}
\showURL{%
\tempurl}


\bibitem[\protect\citeauthoryear{Gerasimou, Eniser, Sen, and Cakan}{Gerasimou
  et~al\mbox{.}}{2020}]%
        {GerasimouICSE20}
\bibfield{author}{\bibinfo{person}{Simos Gerasimou},
  \bibinfo{person}{Hasan~Ferit Eniser}, \bibinfo{person}{Alper Sen}, {and}
  \bibinfo{person}{Alper Cakan}.} \bibinfo{year}{2020}\natexlab{}.
\newblock \showarticletitle{Importance-driven deep learning system testing}. In
  \bibinfo{booktitle}{\emph{{ICSE} '20: 42nd International Conference on
  Software Engineering, Seoul, South Korea, 27 June - 19 July, 2020}},
  \bibfield{editor}{\bibinfo{person}{Gregg Rothermel} {and}
  \bibinfo{person}{Doo{-}Hwan Bae}} (Eds.). \bibinfo{publisher}{{ACM}},
  \bibinfo{pages}{702--713}.
\newblock


\bibitem[\protect\citeauthoryear{Ghanta, Khermosh, Subramanian, Sridhar,
  Sundararaman, Arteaga, Luo, Roselli, Das, and Talagala}{Ghanta
  et~al\mbox{.}}{2018}]%
        {ghanta2018systems}
\bibfield{author}{\bibinfo{person}{Sindhu Ghanta}, \bibinfo{person}{Lior
  Khermosh}, \bibinfo{person}{Sriram Subramanian}, \bibinfo{person}{Vinay
  Sridhar}, \bibinfo{person}{Swaminathan Sundararaman},
  \bibinfo{person}{Dulcardo Arteaga}, \bibinfo{person}{Qianmei Luo},
  \bibinfo{person}{Drew Roselli}, \bibinfo{person}{Dhananjoy Das}, {and}
  \bibinfo{person}{Nisha Talagala}.} \bibinfo{year}{2018}\natexlab{}.
\newblock \showarticletitle{A systems perspective to reproducibility in
  production machine learning domain}.
\newblock  (\bibinfo{year}{2018}).
\newblock


\bibitem[\protect\citeauthoryear{Goldberg}{Goldberg}{1991}]%
        {GoldbergCS91}
\bibfield{author}{\bibinfo{person}{David Goldberg}.}
  \bibinfo{year}{1991}\natexlab{}.
\newblock \showarticletitle{What Every Computer Scientist Should Know About
  Floating-Point Arithmetic}.
\newblock \bibinfo{journal}{\emph{{ACM} Comput. Surv.}} \bibinfo{volume}{23},
  \bibinfo{number}{1} (\bibinfo{year}{1991}), \bibinfo{pages}{5--48}.
\newblock


\bibitem[\protect\citeauthoryear{Grigorescu, Trasnea, Cocias, and
  Macesanu}{Grigorescu et~al\mbox{.}}{2020}]%
        {GrigorescuJFR20}
\bibfield{author}{\bibinfo{person}{Sorin~Mihai Grigorescu},
  \bibinfo{person}{Bogdan Trasnea}, \bibinfo{person}{Tiberiu~T. Cocias}, {and}
  \bibinfo{person}{Gigel Macesanu}.} \bibinfo{year}{2020}\natexlab{}.
\newblock \showarticletitle{A survey of deep learning techniques for autonomous
  driving}.
\newblock \bibinfo{journal}{\emph{J. Field Robotics}} \bibinfo{volume}{37},
  \bibinfo{number}{3} (\bibinfo{year}{2020}), \bibinfo{pages}{362--386}.
\newblock


\bibitem[\protect\citeauthoryear{Gu, Xu, Lu, Zhou, and Wang}{Gu
  et~al\mbox{.}}{2021}]%
        {GuIFIP21}
\bibfield{author}{\bibinfo{person}{Jiazhen Gu}, \bibinfo{person}{Huanlin Xu},
  \bibinfo{person}{Haochuan Lu}, \bibinfo{person}{Yangfan Zhou}, {and}
  \bibinfo{person}{Xin Wang}.} \bibinfo{year}{2021}\natexlab{}.
\newblock \showarticletitle{Detecting Deep Neural Network Defects with Data
  Flow Analysis}. In \bibinfo{booktitle}{\emph{51st Annual {IEEE/IFIP}
  International Conference on Dependable Systems and Networks Workshops, {DSN}
  Workshops 2021, Taipei, Taiwan, June 21-24, 2021}}.
\newblock


\bibitem[\protect\citeauthoryear{Gu, Zhang, and Kim}{Gu et~al\mbox{.}}{2018}]%
        {GuICSE018}
\bibfield{author}{\bibinfo{person}{Xiaodong Gu}, \bibinfo{person}{Hongyu
  Zhang}, {and} \bibinfo{person}{Sunghun Kim}.}
  \bibinfo{year}{2018}\natexlab{}.
\newblock \showarticletitle{Deep code search}. In
  \bibinfo{booktitle}{\emph{Proceedings of the 40th International Conference on
  Software Engineering, {ICSE} 2018, Gothenburg, Sweden, May 27 - June 03,
  2018}}, \bibfield{editor}{\bibinfo{person}{Michel Chaudron},
  \bibinfo{person}{Ivica Crnkovic}, \bibinfo{person}{Marsha Chechik}, {and}
  \bibinfo{person}{Mark Harman}} (Eds.). \bibinfo{publisher}{{ACM}},
  \bibinfo{pages}{933--944}.
\newblock
\urldef\tempurl%
\url{https://doi.org/10.1145/3180155.3180167}
\showDOI{\tempurl}


\bibitem[\protect\citeauthoryear{Gundersen and Kjensmo}{Gundersen and
  Kjensmo}{[n.d.]}]%
        {GundersenAAAI18}
\bibfield{author}{\bibinfo{person}{Odd~Erik Gundersen} {and}
  \bibinfo{person}{Sigbj{\o}rn Kjensmo}.} \bibinfo{year}{[n.d.]}\natexlab{}.
\newblock \showarticletitle{State of the Art: Reproducibility in Artificial
  Intelligence}. In \bibinfo{booktitle}{\emph{Proceedings of the Thirty-Second
  {AAAI} Conference on Artificial Intelligence, (AAAI-18)}}.
\newblock


\bibitem[\protect\citeauthoryear{Guo, Chen, Xie, Ma, Hu, Liu, Liu, Zhao, and
  Li}{Guo et~al\mbox{.}}{2019}]%
        {GuoASE19}
\bibfield{author}{\bibinfo{person}{Qianyu Guo}, \bibinfo{person}{Sen Chen},
  \bibinfo{person}{Xiaofei Xie}, \bibinfo{person}{Lei Ma},
  \bibinfo{person}{Qiang Hu}, \bibinfo{person}{Hongtao Liu},
  \bibinfo{person}{Yang Liu}, \bibinfo{person}{Jianjun Zhao}, {and}
  \bibinfo{person}{Xiaohong Li}.} \bibinfo{year}{2019}\natexlab{}.
\newblock \showarticletitle{An Empirical Study Towards Characterizing Deep
  Learning Development and Deployment Across Different Frameworks and
  Platforms}. In \bibinfo{booktitle}{\emph{34th {IEEE/ACM} International
  Conference on Automated Software Engineering, {ASE} 2019, San Diego, CA, USA,
  November 11-15, 2019}}.
\newblock


\bibitem[\protect\citeauthoryear{He, Zhang, Ren, and Sun}{He
  et~al\mbox{.}}{2016}]%
        {HeCVPR16}
\bibfield{author}{\bibinfo{person}{Kaiming He}, \bibinfo{person}{Xiangyu
  Zhang}, \bibinfo{person}{Shaoqing Ren}, {and} \bibinfo{person}{Jian Sun}.}
  \bibinfo{year}{2016}\natexlab{}.
\newblock \showarticletitle{Deep Residual Learning for Image Recognition}. In
  \bibinfo{booktitle}{\emph{2016 {IEEE} Conference on Computer Vision and
  Pattern Recognition, {CVPR} 2016, Las Vegas, NV, USA, June 27-30, 2016}}.
  \bibinfo{publisher}{{IEEE} Computer Society}, \bibinfo{pages}{770--778}.
\newblock


\bibitem[\protect\citeauthoryear{Henderson, Islam, Bachman, Pineau, Precup, and
  Meger}{Henderson et~al\mbox{.}}{2018}]%
        {HendersonAAAI18}
\bibfield{author}{\bibinfo{person}{Peter Henderson}, \bibinfo{person}{Riashat
  Islam}, \bibinfo{person}{Philip Bachman}, \bibinfo{person}{Joelle Pineau},
  \bibinfo{person}{Doina Precup}, {and} \bibinfo{person}{David Meger}.}
  \bibinfo{year}{2018}\natexlab{}.
\newblock \showarticletitle{Deep Reinforcement Learning That Matters}. In
  \bibinfo{booktitle}{\emph{Proceedings of the Thirty-Second {AAAI} Conference
  on Artificial Intelligence, (AAAI-18)}}. \bibinfo{publisher}{{AAAI} Press}.
\newblock


\bibitem[\protect\citeauthoryear{Hutson}{Hutson}{2018}]%
        {HutsonScience18}
\bibfield{author}{\bibinfo{person}{Matthew Hutson}.}
  \bibinfo{year}{2018}\natexlab{}.
\newblock \showarticletitle{Artificial intelligence faces reproducibility
  crisis}.
\newblock \bibinfo{journal}{\emph{Science (New York, N.Y.)}}
  \bibinfo{volume}{359} (\bibinfo{date}{02} \bibinfo{year}{2018}),
  \bibinfo{pages}{725--726}.
\newblock
\urldef\tempurl%
\url{https://doi.org/10.1126/science.359.6377.725}
\showDOI{\tempurl}


\bibitem[\protect\citeauthoryear{Hutter, Kotthoff, and Vanschoren}{Hutter
  et~al\mbox{.}}{2019}]%
        {HutterAutoML2019}
\bibfield{editor}{\bibinfo{person}{Frank Hutter}, \bibinfo{person}{Lars
  Kotthoff}, {and} \bibinfo{person}{Joaquin Vanschoren}} (Eds.).
  \bibinfo{year}{2019}\natexlab{}.
\newblock \bibinfo{booktitle}{\emph{Automated Machine Learning - Methods,
  Systems, Challenges}}.
\newblock \bibinfo{publisher}{Springer}.
\newblock
\showISBNx{978-3-030-05317-8}
\urldef\tempurl%
\url{https://doi.org/10.1007/978-3-030-05318-5}
\showDOI{\tempurl}


\bibitem[\protect\citeauthoryear{Idowu, Str{\"{u}}ber, and Berger}{Idowu
  et~al\mbox{.}}{2021}]%
        {IdowuSEIP21}
\bibfield{author}{\bibinfo{person}{Samuel Idowu}, \bibinfo{person}{Daniel
  Str{\"{u}}ber}, {and} \bibinfo{person}{Thorsten Berger}.}
  \bibinfo{year}{2021}\natexlab{}.
\newblock \showarticletitle{Asset Management in Machine Learning: {A} Survey}.
  In \bibinfo{booktitle}{\emph{43rd {IEEE/ACM} International Conference on
  Software Engineering: Software Engineering in Practice, {ICSE} {(SEIP)} 2021,
  Madrid, Spain, May 25-28, 2021}}. \bibinfo{publisher}{{IEEE}},
  \bibinfo{pages}{51--60}.
\newblock


\bibitem[\protect\citeauthoryear{Isdahl and Gundersen}{Isdahl and
  Gundersen}{2019}]%
        {isdahl2019out}
\bibfield{author}{\bibinfo{person}{Richard Isdahl} {and}
  \bibinfo{person}{Odd~Erik Gundersen}.} \bibinfo{year}{2019}\natexlab{}.
\newblock \showarticletitle{Out-of-the-box reproducibility: A survey of machine
  learning platforms}. In \bibinfo{booktitle}{\emph{2019 15th international
  conference on eScience (eScience)}}. IEEE, \bibinfo{pages}{86--95}.
\newblock


\bibitem[\protect\citeauthoryear{Jooybar, Fung, O'Connor, Devietti, and
  Aamodt}{Jooybar et~al\mbox{.}}{2013}]%
        {JooybarASPLOS13}
\bibfield{author}{\bibinfo{person}{Hadi Jooybar}, \bibinfo{person}{Wilson W.~L.
  Fung}, \bibinfo{person}{Mike O'Connor}, \bibinfo{person}{Joseph Devietti},
  {and} \bibinfo{person}{Tor~M. Aamodt}.} \bibinfo{year}{2013}\natexlab{}.
\newblock \showarticletitle{GPUDet: a deterministic {GPU} architecture}. In
  \bibinfo{booktitle}{\emph{Architectural Support for Programming Languages and
  Operating Systems, {ASPLOS} 2013, Houston, TX, USA, March 16-20, 2013}}.
  \bibinfo{publisher}{{ACM}}, \bibinfo{pages}{1--12}.
\newblock
\urldef\tempurl%
\url{https://doi.org/10.1145/2451116.2451118}
\showDOI{\tempurl}


\bibitem[\protect\citeauthoryear{LeCun, Bengio, and Hinton}{LeCun
  et~al\mbox{.}}{2015}]%
        {LeCunNature15}
\bibfield{author}{\bibinfo{person}{Yann LeCun}, \bibinfo{person}{Yoshua
  Bengio}, {and} \bibinfo{person}{Geoffrey~E. Hinton}.}
  \bibinfo{year}{2015}\natexlab{}.
\newblock \showarticletitle{Deep learning}.
\newblock \bibinfo{journal}{\emph{Nat.}} (\bibinfo{year}{2015}).
\newblock


\bibitem[\protect\citeauthoryear{Lecun, Bottou, Bengio, and Haffner}{Lecun
  et~al\mbox{.}}{1998}]%
        {LecunIEEE98}
\bibfield{author}{\bibinfo{person}{Y. Lecun}, \bibinfo{person}{L. Bottou},
  \bibinfo{person}{Y. Bengio}, {and} \bibinfo{person}{P. Haffner}.}
  \bibinfo{year}{1998}\natexlab{}.
\newblock \showarticletitle{Gradient-based learning applied to document
  recognition}.
\newblock \bibinfo{journal}{\emph{Proc. IEEE}} (\bibinfo{year}{1998}).
\newblock


\bibitem[\protect\citeauthoryear{Lee, Jackson, Madams, Troisi, and Jones}{Lee
  et~al\mbox{.}}{2019}]%
        {LeeRML19}
\bibfield{author}{\bibinfo{person}{Brian Lee}, \bibinfo{person}{Andrew
  Jackson}, \bibinfo{person}{Tom Madams}, \bibinfo{person}{Seth Troisi}, {and}
  \bibinfo{person}{Derek Jones}.} \bibinfo{year}{2019}\natexlab{}.
\newblock \showarticletitle{Minigo: {A} Case Study in Reproducing Reinforcement
  Learning Research}. In \bibinfo{booktitle}{\emph{Reproducibility in Machine
  Learning, {ICLR} 2019 Workshop, New Orleans, Louisiana, United States, May 6,
  2019}}. \bibinfo{publisher}{OpenReview.net}.
\newblock


\bibitem[\protect\citeauthoryear{Li}{Li}{2017}]%
        {bigballon2017cifar10cnn}
\bibfield{author}{\bibinfo{person}{Wei Li}.} \bibinfo{year}{2017}\natexlab{}.
\newblock \bibinfo{title}{cifar-10-cnn: Play deep learning with CIFAR
  datasets}.
\newblock
  \bibinfo{howpublished}{\url{https://github.com/BIGBALLON/cifar-10-cnn}}.
\newblock


\bibitem[\protect\citeauthoryear{Liu, Gao, Xia, Lo, Grundy, and Yang}{Liu
  et~al\mbox{.}}{2020}]%
        {LiuTOSEM20}
\bibfield{author}{\bibinfo{person}{Chao Liu}, \bibinfo{person}{Cuiyun Gao},
  \bibinfo{person}{Xin Xia}, \bibinfo{person}{David Lo},
  \bibinfo{person}{John~C. Grundy}, {and} \bibinfo{person}{Xiaohu Yang}.}
  \bibinfo{year}{2020}\natexlab{}.
\newblock \showarticletitle{On the Replicability and Reproducibility of Deep
  Learning in Software Engineering}.
\newblock \bibinfo{journal}{\emph{CoRR}}  \bibinfo{volume}{abs/2006.14244}
  (\bibinfo{year}{2020}).
\newblock
\showeprint[arxiv]{2006.14244}
\urldef\tempurl%
\url{https://arxiv.org/abs/2006.14244}
\showURL{%
\tempurl}


\bibitem[\protect\citeauthoryear{Ma, Juefei{-}Xu, Zhang, Sun, Xue, Li, Chen,
  Su, Li, Liu, Zhao, and Wang}{Ma et~al\mbox{.}}{2018a}]%
        {MaDeepGaugeASE18}
\bibfield{author}{\bibinfo{person}{Lei Ma}, \bibinfo{person}{Felix
  Juefei{-}Xu}, \bibinfo{person}{Fuyuan Zhang}, \bibinfo{person}{Jiyuan Sun},
  \bibinfo{person}{Minhui Xue}, \bibinfo{person}{Bo Li},
  \bibinfo{person}{Chunyang Chen}, \bibinfo{person}{Ting Su},
  \bibinfo{person}{Li Li}, \bibinfo{person}{Yang Liu}, \bibinfo{person}{Jianjun
  Zhao}, {and} \bibinfo{person}{Yadong Wang}.}
  \bibinfo{year}{2018}\natexlab{a}.
\newblock \showarticletitle{DeepGauge: multi-granularity testing criteria for
  deep learning systems}. In \bibinfo{booktitle}{\emph{Proceedings of the 33rd
  {ACM/IEEE} International Conference on Automated Software Engineering, {ASE}
  2018, Montpellier, France, September 3-7, 2018}},
  \bibfield{editor}{\bibinfo{person}{Marianne Huchard},
  \bibinfo{person}{Christian K{\"{a}}stner}, {and} \bibinfo{person}{Gordon
  Fraser}} (Eds.). \bibinfo{publisher}{{ACM}}, \bibinfo{pages}{120--131}.
\newblock


\bibitem[\protect\citeauthoryear{Ma, Liu, Lee, Zhang, and Grama}{Ma
  et~al\mbox{.}}{2018b}]%
        {MaFSE18}
\bibfield{author}{\bibinfo{person}{Shiqing Ma}, \bibinfo{person}{Yingqi Liu},
  \bibinfo{person}{Wen{-}Chuan Lee}, \bibinfo{person}{Xiangyu Zhang}, {and}
  \bibinfo{person}{Ananth Grama}.} \bibinfo{year}{2018}\natexlab{b}.
\newblock \showarticletitle{{MODE:} automated neural network model debugging
  via state differential analysis and input selection}. In
  \bibinfo{booktitle}{\emph{Proceedings of the 2018 {ACM} Joint Meeting on
  European Software Engineering Conference and Symposium on the Foundations of
  Software Engineering, {ESEC/SIGSOFT} {FSE} 2018, Lake Buena Vista, FL, USA,
  November 04-09, 2018}}, \bibfield{editor}{\bibinfo{person}{Gary~T. Leavens},
  \bibinfo{person}{Alessandro Garcia}, {and} \bibinfo{person}{Corina~S.
  Pasareanu}} (Eds.).
\newblock


\bibitem[\protect\citeauthoryear{Mitchell, Wu, Zaldivar, Barnes, Vasserman,
  Hutchinson, Spitzer, Raji, and Gebru}{Mitchell et~al\mbox{.}}{2019}]%
        {MitchellFAT19}
\bibfield{author}{\bibinfo{person}{Margaret Mitchell}, \bibinfo{person}{Simone
  Wu}, \bibinfo{person}{Andrew Zaldivar}, \bibinfo{person}{Parker Barnes},
  \bibinfo{person}{Lucy Vasserman}, \bibinfo{person}{Ben Hutchinson},
  \bibinfo{person}{Elena Spitzer}, \bibinfo{person}{Inioluwa~Deborah Raji},
  {and} \bibinfo{person}{Timnit Gebru}.} \bibinfo{year}{2019}\natexlab{}.
\newblock \showarticletitle{Model Cards for Model Reporting}. In
  \bibinfo{booktitle}{\emph{Proceedings of the Conference on Fairness,
  Accountability, and Transparency, FAT* 2019, Atlanta, GA, USA, January 29-31,
  2019}}, \bibfield{editor}{\bibinfo{person}{danah boyd} {and}
  \bibinfo{person}{Jamie~H. Morgenstern}} (Eds.). \bibinfo{publisher}{{ACM}},
  \bibinfo{pages}{220--229}.
\newblock


\bibitem[\protect\citeauthoryear{Parnas}{Parnas}{2017}]%
        {ParnasACM17}
\bibfield{author}{\bibinfo{person}{David~Lorge Parnas}.}
  \bibinfo{year}{2017}\natexlab{}.
\newblock \showarticletitle{The real risks of artificial intelligence}.
\newblock \bibinfo{journal}{\emph{Commun. {ACM}}} \bibinfo{volume}{60},
  \bibinfo{number}{10} (\bibinfo{year}{2017}), \bibinfo{pages}{27--31}.
\newblock
\urldef\tempurl%
\url{https://doi.org/10.1145/3132724}
\showDOI{\tempurl}


\bibitem[\protect\citeauthoryear{Paszke, Gross, Massa, Lerer, Bradbury, Chanan,
  Killeen, Lin, Gimelshein, Antiga, Desmaison, K{\"{o}}pf, Yang, DeVito,
  Raison, Tejani, Chilamkurthy, Steiner, Fang, Bai, and Chintala}{Paszke
  et~al\mbox{.}}{2019}]%
        {PaszkeNIPS19}
\bibfield{author}{\bibinfo{person}{Adam Paszke}, \bibinfo{person}{Sam Gross},
  \bibinfo{person}{Francisco Massa}, \bibinfo{person}{Adam Lerer},
  \bibinfo{person}{James Bradbury}, \bibinfo{person}{Gregory Chanan},
  \bibinfo{person}{Trevor Killeen}, \bibinfo{person}{Zeming Lin},
  \bibinfo{person}{Natalia Gimelshein}, \bibinfo{person}{Luca Antiga},
  \bibinfo{person}{Alban Desmaison}, \bibinfo{person}{Andreas K{\"{o}}pf},
  \bibinfo{person}{Edward Yang}, \bibinfo{person}{Zachary DeVito},
  \bibinfo{person}{Martin Raison}, \bibinfo{person}{Alykhan Tejani},
  \bibinfo{person}{Sasank Chilamkurthy}, \bibinfo{person}{Benoit Steiner},
  \bibinfo{person}{Lu Fang}, \bibinfo{person}{Junjie Bai}, {and}
  \bibinfo{person}{Soumith Chintala}.} \bibinfo{year}{2019}\natexlab{}.
\newblock \showarticletitle{PyTorch: An Imperative Style, High-Performance Deep
  Learning Library}. In \bibinfo{booktitle}{\emph{Advances in Neural
  Information Processing Systems 32: Annual Conference on Neural Information
  Processing Systems 2019, NeurIPS 2019, December 8-14, 2019, Vancouver, BC,
  Canada}}, \bibfield{editor}{\bibinfo{person}{Hanna~M. Wallach},
  \bibinfo{person}{Hugo Larochelle}, \bibinfo{person}{Alina Beygelzimer},
  \bibinfo{person}{Florence d'Alch{\'{e}}{-}Buc}, \bibinfo{person}{Emily~B.
  Fox}, {and} \bibinfo{person}{Roman Garnett}} (Eds.).
  \bibinfo{pages}{8024--8035}.
\newblock


\bibitem[\protect\citeauthoryear{Pham, Qian, Wang, Lutellier, Rosenthal, Tan,
  Yu, and Nagappan}{Pham et~al\mbox{.}}{[n.d.]}]%
        {PhamASE20}
\bibfield{author}{\bibinfo{person}{Hung~Viet Pham}, \bibinfo{person}{Shangshu
  Qian}, \bibinfo{person}{Jiannan Wang}, \bibinfo{person}{Thibaud Lutellier},
  \bibinfo{person}{Jonathan Rosenthal}, \bibinfo{person}{Lin Tan},
  \bibinfo{person}{Yaoliang Yu}, {and} \bibinfo{person}{Nachiappan Nagappan}.}
  \bibinfo{year}{[n.d.]}\natexlab{}.
\newblock \showarticletitle{Problems and Opportunities in Training Deep
  Learning Software Systems: An Analysis of Variance}. In
  \bibinfo{booktitle}{\emph{35th {IEEE/ACM} International Conference on
  Automated Software Engineering, {ASE} 2020, Melbourne, Australia, September
  21-25, 2020}}.
\newblock


\bibitem[\protect\citeauthoryear{Pineau, Vincent-Lamarre, Sinha, Larivière,
  Beygelzimer, d'Alché Buc, Fox, and Larochelle}{Pineau et~al\mbox{.}}{2020}]%
        {pineau2020improving}
\bibfield{author}{\bibinfo{person}{Joelle Pineau}, \bibinfo{person}{Philippe
  Vincent-Lamarre}, \bibinfo{person}{Koustuv Sinha}, \bibinfo{person}{Vincent
  Larivière}, \bibinfo{person}{Alina Beygelzimer}, \bibinfo{person}{Florence
  d'Alché Buc}, \bibinfo{person}{Emily Fox}, {and} \bibinfo{person}{Hugo
  Larochelle}.} \bibinfo{year}{2020}\natexlab{}.
\newblock \bibinfo{title}{Improving Reproducibility in Machine Learning
  Research (A Report from the NeurIPS 2019 Reproducibility Program)}.
\newblock
\newblock
\showeprint[arxiv]{2003.12206}~[cs.LG]


\bibitem[\protect\citeauthoryear{Raff}{Raff}{2019}]%
        {RaffNIPS19}
\bibfield{author}{\bibinfo{person}{Edward Raff}.}
  \bibinfo{year}{2019}\natexlab{}.
\newblock \showarticletitle{A Step Toward Quantifying Independently
  Reproducible Machine Learning Research}. In
  \bibinfo{booktitle}{\emph{Advances in Neural Information Processing Systems
  32: Annual Conference on Neural Information Processing Systems 2019, NeurIPS
  2019, December 8-14, 2019, Vancouver, BC, Canada}},
  \bibfield{editor}{\bibinfo{person}{Hanna~M. Wallach}, \bibinfo{person}{Hugo
  Larochelle}, \bibinfo{person}{Alina Beygelzimer}, \bibinfo{person}{Florence
  d'Alch{\'{e}}{-}Buc}, \bibinfo{person}{Emily~B. Fox}, {and}
  \bibinfo{person}{Roman Garnett}} (Eds.).
\newblock


\bibitem[\protect\citeauthoryear{Romano, Kromrey, Coraggio, and
  Skowronek}{Romano et~al\mbox{.}}{2006}]%
        {romano2006appropriate}
\bibfield{author}{\bibinfo{person}{Jeanine Romano}, \bibinfo{person}{Jeffrey~D
  Kromrey}, \bibinfo{person}{Jesse Coraggio}, {and} \bibinfo{person}{Jeff
  Skowronek}.} \bibinfo{year}{2006}\natexlab{}.
\newblock \showarticletitle{Appropriate statistics for ordinal level data:
  Should we really be using t-test and Cohen’sd for evaluating group
  differences on the NSSE and other surveys}. In
  \bibinfo{booktitle}{\emph{annual meeting of the Florida Association of
  Institutional Research}}, Vol.~\bibinfo{volume}{13}.
\newblock


\bibitem[\protect\citeauthoryear{Salmon, Moraes, Dror, and Shaw}{Salmon
  et~al\mbox{.}}{2011}]%
        {SalmonSC11}
\bibfield{author}{\bibinfo{person}{John~K. Salmon}, \bibinfo{person}{Mark~A.
  Moraes}, \bibinfo{person}{Ron~O. Dror}, {and} \bibinfo{person}{David~E.
  Shaw}.} \bibinfo{year}{2011}\natexlab{}.
\newblock \showarticletitle{Parallel random numbers: as easy as 1, 2, 3}. In
  \bibinfo{booktitle}{\emph{Conference on High Performance Computing
  Networking, Storage and Analysis, {SC} 2011, Seattle, WA, USA, November
  12-18, 2011}}. \bibinfo{publisher}{{ACM}}, \bibinfo{pages}{16:1--16:12}.
\newblock
\urldef\tempurl%
\url{https://doi.org/10.1145/2063384.2063405}
\showDOI{\tempurl}


\bibitem[\protect\citeauthoryear{Scardapane and Wang}{Scardapane and
  Wang}{2017}]%
        {ScardapaneWI17}
\bibfield{author}{\bibinfo{person}{Simone Scardapane} {and}
  \bibinfo{person}{Dianhui Wang}.} \bibinfo{year}{2017}\natexlab{}.
\newblock \showarticletitle{Randomness in neural networks: an overview}.
\newblock \bibinfo{journal}{\emph{Wiley Interdiscip. Rev. Data Min. Knowl.
  Discov.}} \bibinfo{volume}{7}, \bibinfo{number}{2} (\bibinfo{year}{2017}).
\newblock


\bibitem[\protect\citeauthoryear{Scheuner, Cito, Leitner, and Gall}{Scheuner
  et~al\mbox{.}}{2015}]%
        {ScheunerWWW15}
\bibfield{author}{\bibinfo{person}{Joel Scheuner},
  \bibinfo{person}{J{\"{u}}rgen Cito}, \bibinfo{person}{Philipp Leitner}, {and}
  \bibinfo{person}{Harald~C. Gall}.} \bibinfo{year}{2015}\natexlab{}.
\newblock \showarticletitle{Cloud WorkBench: Benchmarking IaaS Providers based
  on Infrastructure-as-Code}. In \bibinfo{booktitle}{\emph{Proceedings of the
  24th International Conference on World Wide Web Companion, {WWW} 2015,
  Florence, Italy, May 18-22, 2015 - Companion Volume}}.
  \bibinfo{publisher}{{ACM}}, \bibinfo{pages}{239--242}.
\newblock
\urldef\tempurl%
\url{https://doi.org/10.1145/2740908.2742833}
\showDOI{\tempurl}


\bibitem[\protect\citeauthoryear{Sugimura and Hartl}{Sugimura and
  Hartl}{2018a}]%
        {sugimura2018building}
\bibfield{author}{\bibinfo{person}{Peter Sugimura} {and}
  \bibinfo{person}{Florian Hartl}.} \bibinfo{year}{2018}\natexlab{a}.
\newblock \showarticletitle{Building a reproducible machine learning pipeline}.
\newblock \bibinfo{journal}{\emph{arXiv preprint arXiv:1810.04570}}
  (\bibinfo{year}{2018}).
\newblock


\bibitem[\protect\citeauthoryear{Sugimura and Hartl}{Sugimura and
  Hartl}{2018b}]%
        {SugimuraArxiv18}
\bibfield{author}{\bibinfo{person}{Peter Sugimura} {and}
  \bibinfo{person}{Florian Hartl}.} \bibinfo{year}{2018}\natexlab{b}.
\newblock \showarticletitle{Building a Reproducible Machine Learning Pipeline}.
\newblock \bibinfo{journal}{\emph{CoRR}}  \bibinfo{volume}{abs/1810.04570}
  (\bibinfo{year}{2018}).
\newblock
\showeprint[arxiv]{1810.04570}
\urldef\tempurl%
\url{http://arxiv.org/abs/1810.04570}
\showURL{%
\tempurl}


\bibitem[\protect\citeauthoryear{Tatman, VanderPlas, and Dane}{Tatman
  et~al\mbox{.}}{2018}]%
        {tatman2018practical}
\bibfield{author}{\bibinfo{person}{Rachael Tatman}, \bibinfo{person}{Jake
  VanderPlas}, {and} \bibinfo{person}{Sohier Dane}.}
  \bibinfo{year}{2018}\natexlab{}.
\newblock \showarticletitle{A practical taxonomy of reproducibility for machine
  learning research}.
\newblock  (\bibinfo{year}{2018}).
\newblock


\bibitem[\protect\citeauthoryear{Vicente-Saez and
  Martinez-Fuentes}{Vicente-Saez and Martinez-Fuentes}{2018}]%
        {vicente2018open}
\bibfield{author}{\bibinfo{person}{Ruben Vicente-Saez} {and}
  \bibinfo{person}{Clara Martinez-Fuentes}.} \bibinfo{year}{2018}\natexlab{}.
\newblock \showarticletitle{Open Science now: A systematic literature review
  for an integrated definition}.
\newblock \bibinfo{journal}{\emph{Journal of business research}}
  \bibinfo{volume}{88} (\bibinfo{year}{2018}), \bibinfo{pages}{428--436}.
\newblock


\bibitem[\protect\citeauthoryear{Woelfle, Olliaro, and Todd}{Woelfle
  et~al\mbox{.}}{2011}]%
        {woelfle2011open}
\bibfield{author}{\bibinfo{person}{Michael Woelfle}, \bibinfo{person}{Piero
  Olliaro}, {and} \bibinfo{person}{Matthew~H Todd}.}
  \bibinfo{year}{2011}\natexlab{}.
\newblock \showarticletitle{Open science is a research accelerator}.
\newblock \bibinfo{journal}{\emph{Nature chemistry}} \bibinfo{volume}{3},
  \bibinfo{number}{10} (\bibinfo{year}{2011}), \bibinfo{pages}{745--748}.
\newblock


\bibitem[\protect\citeauthoryear{Yanko}{Yanko}{2021}]%
        {SBOM:2020}
\bibfield{author}{\bibinfo{person}{Curtis Yanko}.} \bibinfo{year}{2021
  (accessed August, 2021)}\natexlab{}.
\newblock \bibinfo{booktitle}{\emph{Using a Software Bill of Materials (SBOM)
  is Going Mainstream}}.
\newblock
\urldef\tempurl%
\url{https://blog.sonatype.com/software-bill-of-materials-going-mainstream}
\showURL{%
\tempurl}


\bibitem[\protect\citeauthoryear{Zagoruyko and Komodakis}{Zagoruyko and
  Komodakis}{2016}]%
        {WRN16}
\bibfield{author}{\bibinfo{person}{Sergey Zagoruyko} {and}
  \bibinfo{person}{Nikos Komodakis}.} \bibinfo{year}{2016}\natexlab{}.
\newblock \showarticletitle{Wide Residual Networks}. In
  \bibinfo{booktitle}{\emph{Proceedings of the British Machine Vision
  Conference 2016, {BMVC} 2016, York, UK, September 19-22, 2016}}.
  \bibinfo{publisher}{{BMVA} Press}.
\newblock
\urldef\tempurl%
\url{http://www.bmva.org/bmvc/2016/papers/paper087/index.html}
\showURL{%
\tempurl}


\end{thebibliography}

\end{document}